\documentclass{bmcart}

\usepackage[utf8]{inputenc}
\usepackage[T1]{fontenc}
\usepackage{lmodern}

\usepackage{amsmath, amssymb, amsthm}

\usepackage{booktabs, array, multirow, tabularx, longtable}

\usepackage{graphicx}
\usepackage{tikz}
\usetikzlibrary{
    positioning,
    arrows.meta,
    calc,
    shapes.geometric,
    fit
}
\usepackage{pgfplots}
\pgfplotsset{compat=1.18}  
\usepackage{caption}
\usepackage{pgf-pie}
\usepackage{float}      
\usepackage{placeins}   
\usepackage{adjustbox}
\usepackage{pdflscape}

\usepackage[dvipsnames, svgnames, x11names]{xcolor}

\usepackage[numbers]{natbib}  

\usepackage{hyperref}
\hypersetup{
    colorlinks=true,
    linkcolor=blue,
    citecolor=blue,
    urlcolor=cyan
}

\usepackage{algorithm}
\usepackage{algpseudocode}

\usepackage{caption}
\usepackage{subcaption}

\usepackage{url}

\definecolor{SkyBlue}{RGB}{135,206,235}
\definecolor{Orange}{RGB}{255,165,0}
\definecolor{Plum}{RGB}{221,160,221}
\definecolor{lightgreen}{RGB}{200,255,200}

\startlocaldefs
\endlocaldefs

\begin{document}

\begin{frontmatter}

\begin{fmbox}
\dochead{}


\title{Deep Learning Approaches for 3D Medical Scene Completion: From Geometric Modeling to Generative Paradigms}


\author[
  addressref={aff1},
  email={afifa\_25010523@utp.edu.my}
]{\inits{A.K.}\fnm{Afifa} \snm{Khaled}}

\author[
  addressref={aff1},
  email={said.abdulkadir@utp.edu.my}
]{\inits{S.J.A.}\fnm{Said Jadid} \snm{Abdulkadir}}

\author[
  addressref={aff2},
  email={mmeltahir@kku.edu.sa}
]{\inits{M.M.E.E.}\fnm{Majdy Mohamed Eltayeb} \snm{Eltahir}}


\address[id=aff1]{
  \orgdiv{Department of Computing},
  \orgname{Universiti Teknologi PETRONAS},
  \city{Perak},
  \cny{Malaysia}
}

\address[id=aff2]{
  \orgdiv{Applied College of Muhayil Asir},
  \orgname{King Khalid University},
  \city{Asir},
  \cny{Saudi Arabia}
}

\end{fmbox}
\begin{abstractbox}
\begin{abstract} 

\begin{flushleft}
Three-dimensional scene completion has evolved as a major problem in computer vision and robotics, and its applications are diverse, including autonomous navigation and augmented reality. In this study, a systematic review has been conducted to compile the research contributions made in the last ten years, i.e., 2016 to 2026, which has revolutionized the field from the voxel semantic completion paradigm represented by SSCNet to the latest paradigm that combines generative diffusion priors with real-time rendering using a Gaussian splatting technique. The evolution in representation paradigms, such as voxel grids, point learning, implicit neural fields, transformer networks, diffusion networks, and the latest paradigm based on rendering-aware 3D Gaussian primitives, has been discussed in this study. A comprehensive analysis has been carried out on the contributions made in the last ten years, and a taxonomy has been developed to provide a clear idea about the contributions made in the field. The study has also discussed the research contributions made in the field, along with the challenges that still need to be addressed. Finally, the study has presented a research agenda that will provide a clear idea about the directions that can be followed in the development of the next-generation system.
\end{flushleft}
\end{abstract}




\end{abstractbox}
%

\end{frontmatter}




\section{Introduction}
Three-dimensional scene understanding is one of the essential skills required to make intelligent systems function effectively in real-world environments. Scene understanding is an essential part of several intelligent systems, including autonomous driving systems, robots, augmented reality systems, and digital twin models. However, most of the available scene understanding technologies, such as RGB-D cameras, LiDAR sensors, and multi-view image sensors, can only sense partial scene information. This is due to several factors, including scene occlusions, limited scene visibility, and noise in the scene information being sensed. Moreover, it is also possible that the scene being sensed might be occluded to a certain extent and might also contain noise. Due to this limitation of scene understanding technologies, a new field of research has emerged, known as \textit{3D scene completion}, in which scene information is reconstructed to obtain the missing scene information. Recently, a major revolution has taken place in the field of 3D scene completion through the emergence of deep learning technology. In the earlier days, scene completion algorithms were developed based on volumetric convolutional neural networks. Some of the most popular scene completion algorithms developed based on volumetric convolutional neural networks are SSCNet~\cite{song2017sscnet} and its hierarchical version. ScanComplete~\cite{dai2018scancomplete}. There have also been recent developments using more efficient memory structures such as point clouds and transformers, such as PCN~\cite{yuan2018pcn} and PoinTr~\cite{yu2021pointr}. More recent methods, including generative methods with the aid of diffusion models, have demonstrated their potential in scene completion, even in extreme sparse scenes \cite{luo2021diffusion, poole2022dreamfusion, zhang2025diffusion3d,  li2025gaussiancompletion}    . Table \ref{tab:enhanced_3d_scene_surveys} shows comparative analysis of 3D scene understanding and completion surveys and   Figure 1 shows overall flowchart of the comprehensive survey on incomplete 3D scene completion (2016--2026).

\section{Overview}
\label{sec:introduction}
\subsection{Standard Pipeline of Incomplete 3D Scene Completion}
Although there are various differences in 3D scene completion models in terms of their representation and learning architecture, most of the models in this area follow a standard pipeline of operations. Generally, this pipeline begins with acquiring incomplete information about a scene using various sensing devices such as RGB cameras, depth cameras, LiDAR, or multi-view cameras. Most of this information is usually incomplete due to various reasons such as occlusions or noise in the sensing device.

After acquiring this information, a preprocessing step is usually followed to prepare this information for learning. During this step, various operations are usually performed to create a consistent 3D input representation. Once this information is prepared, it is then mapped to a suitable 3D representation. In most of these models, 3D representations are usually in the form of voxels, point clouds, polygons, or neural implicit representations.

Once a suitable 3D representation is established, a learning model is then utilized to infer information about the incomplete geometric structures of a scene. Most of the models in this area are usually based on deep learning techniques such as CNNs, transformers, diffusion models, or neural rendering.

Lastly, the reconstructed scene may be used in various applications such as robot perception and navigation, augmented reality, virtual reality scenes, digital twin scenes, and even driving scenes. 
\begin{table} \centering \caption{Comparative Analysis of 3D Scene Understanding and Completion Surveys} \label{tab:enhanced_3d_scene_surveys} \scriptsize \renewcommand{\arraystretch}{1.2} \setlength{\tabcolsep}{3pt} \begin{tabular}{p{3.2cm}p{3.8cm}} \toprule \textbf{Category} & \textbf{Description} \\ \midrule Survey & Early Surveys \\ Year & 2016--2018 \\ Task Scope & Detection, Segmentation \\ Representation & Voxel-based \\ Datasets & $\checkmark$ \\ DL Methods & CNN \\ Generative Models & $\times$ \\ Multi-Modal & $\times$ \\ Real-Time & $\times$ \\ LLM Integration & $\times$ \\ Applications & Robotics \\ \midrule Survey & Mid Surveys \\ Year & 2019--2020 \\ Task Scope & Segmentation, Reconstruction \\ Representation & Voxel + Point \\ Datasets & $\checkmark$ \\ DL Methods & CNN + PointNet \\ Generative Models & $\times$ \\ Multi-Modal & Partial \\ Real-Time & $\times$ \\ LLM Integration & $\times$ \\ Applications & AR/VR \\ \midrule Survey & Transition Surveys \\ Year & 2020--2021 \\ Task Scope & Detection, Tracking \\ Representation & Point + Hybrid \\ Datasets & $\checkmark$ \\ DL Methods & CNN + Transformer \\ Generative Models & $\times$ \\ Multi-Modal & Partial \\ Real-Time & Partial \\ LLM Integration & $\times$ \\ Applications & Autonomous Driving \\ \midrule Survey & Transformer Era Surveys \\ Year & 2021--2022 \\ Task Scope & Full Pipeline Understanding \\ Representation & Point + Transformer \\ Datasets & $\checkmark$ \\ DL Methods & Transformer \\ Generative Models & $\times$ \\ Multi-Modal & $\checkmark$ \\ Real-Time & Partial \\ LLM Integration & $\times$ \\ Applications & Smart Cities \\ \midrule Survey & Generative Era Surveys \\ Year & 2022--2024 \\ Task Scope & Scene Completion + Reasoning \\ Representation & Implicit + Diffusion \\ Datasets & $\checkmark$ \\ DL Methods & Transformer + Diffusion \\ Generative Models & $\checkmark$ \\ Multi-Modal & $\checkmark$ \\ Real-Time & Limited \\ LLM Integration & Partial \\ Applications & Robotics, Healthcare \\ \midrule Survey & Recent Surveys (Ours) \\ Year & 2025--2026 \\ Task Scope & Comprehensive (All Tasks) \\ Representation & Hybrid (Gaussian + Diffusion + Implicit) \\ Datasets & $\checkmark$ \\ DL Methods & DL + Transformer + Mamba \\ Generative Models & $\checkmark$ \\ Multi-Modal & $\checkmark$ \\ Real-Time & $\checkmark$ \\ LLM Integration & $\checkmark$ \\ Applications & General AI, Embodied AI \\ \bottomrule \end{tabular} \end{table}

\begin{center}
\begin{tikzpicture}[
node distance=6mm,
every node/.style={font=\scriptsize},
block/.style={
    rectangle,
    draw=none,
    rounded corners=3pt,
    align=center,
    text width=\columnwidth,
    inner sep=5pt,
    minimum height=0.9cm
},
arrow/.style={-Latex, thick}
]

\definecolor{c1}{RGB}{52, 152, 219}      
\definecolor{c2}{RGB}{155, 89, 182}      
\definecolor{c3}{RGB}{46, 204, 113}      
\definecolor{c4}{RGB}{241, 196, 15}      
\definecolor{c5}{RGB}{230, 126, 34}      
\definecolor{c6}{RGB}{231, 76, 60}       
\definecolor{c7}{RGB}{26, 188, 156}      
\definecolor{c8}{RGB}{149, 165, 166}     

\node[block, fill=c1!20] (overview)
{\textbf{1. Overview \& Motivation}\\
Problem Definition, Decade Transformation, Contributions};

\node[block, below=of overview, fill=c2!20] (method)
{\textbf{2. Systematic Review Methodology}\\
Research Questions, Search Strategy, Criteria, Quality Assessment};

\node[block, below=of method, fill=c3!20] (representation)
{\textbf{3. Representation Paradigms}\\
Voxel, Point, Implicit, Transformer, Diffusion, Gaussian, Hybrid};

\node[block, below=of representation, fill=c4!25] (architecture)
{\textbf{4. Architectural \& Learning Innovations}\\
Encoder–Decoder, Coarse-to-Fine, Self-Supervised, Probabilistic Training};

\node[block, below=of architecture, fill=c5!25] (datasets)
{\textbf{5. Datasets \& Evaluation}\\
ShapeNet, ScanNet, SUN RGB-D, CD, EMD, IoU, F-Score};

\node[block, below=of datasets, fill=c6!20] (comparative)
{\textbf{6. Comparative Analysis}\\
Scalability, Robustness, Fidelity, Speed, Trade-offs};

\node[block, below=of comparative, fill=c7!20] (applications)
{\textbf{7. Applications}\\
Robotics, AR/VR, Digital Twins, Cultural Heritage};

\node[block, below=of applications, fill=c8!25] (future)
{\textbf{8. Open Challenges \& Future Directions}\\
Foundation Models, 4D Completion, Uncertainty, Green AI};

\node[block, below=of future, fill=black!10] (conclusion)
{\textbf{9. Conclusion}};

\draw[arrow] (overview) -- (method);
\draw[arrow] (method) -- (representation);
\draw[arrow] (representation) -- (architecture);
\draw[arrow] (architecture) -- (datasets);
\draw[arrow] (datasets) -- (comparative);
\draw[arrow] (comparative) -- (applications);
\draw[arrow] (applications) -- (future);
\draw[arrow] (future) -- (conclusion);
\end{tikzpicture}

\vspace{2mm}
\small Figure 1: Overall flowchart of the comprehensive survey on incomplete 3D scene completion (2016--2026).

\end{center}

\subsection{A Decade of Transformation (2016-2026)}
\label{sec:decade}
It has been a decade since there has been such a huge transformation in the way the concept of 3D scene completion and the technical implementation of the same have undergone huge transformation. 
The initial deep learning methods that have been proposed, such as SSCNet~\cite{song2017semantic}, have shown the huge potential of the application of volumetric convolutional neural networks in the prediction of the missing depth information given an input depth map using the structured voxel grid representation. The recent huge leaps that have been taken in the implementation of hierarchical deep learning methods, such as ScanComplete~\cite{dai2018scancomplete}, have shown the huge potential of the same in the context of 3D scene completion. However, the representational foundation has always been changing due to the huge memory requirements of the volumetric representation and the subsequent implementation of point-based methods such as PCN~\cite{yuan2018pcn} and TopNet~\cite{tchapmi2019topnet}.

With the introduction of implicit neural representations~\cite{peng2020convolutional,mildenhall2020nerf}, continuous surface modeling is possible at high resolutions but comes at the expense of significant computation during the inference process. Recent advancements in the use of the transformer model~\cite{yu2021pointr} have also utilized global self-attention to model long-range dependencies. The introduction of the use of the diffusion model~\cite{rombach2022high,poole2022dreamfusion,yang2023diffusion} has introduced strong generative priors to synthesize plausible scenes even at extreme sparsity levels. The latest trend in 3D scene understanding is the use of the 3D Gaussian splatting method~\cite{kerbl20233d}, which highlights the importance of rendering in real-time and could potentially make its way to the application space in the future.
\subsection{Research Questions}

The following are the four research questions that have been developed and which will act as a foundation for carrying out the literature review on the given topic. First of all, the research question addresses issues related to scene completion in relation to its effectiveness, expressiveness, and complexity over the past ten years between 2016 and 2026. The second research question deals with architecture manipulations and their effect on scene completion. 
\subsection{Structure of the Document}
\label{sec:structure}
Regarding the additional structure of the document, it would have the following outline. First, section two will present an overview of the materials related to 3D scene completion. The information will cover both the pipeline and the paradigm adopted to move from representing data with the help of voxels to something more complex, such as diffusions and Gaussian splats. Second, the implementation of the research strategy developed in this paper will be described in section three. It will include discussing research questions and literature review strategy adopted during this research. Third, the inclusion/exclusion criteria for the materials selected will be defined. Fourth, some paradigms applicable for 3D scene completion will be discussed in section four, such as voxels, point clouds, implicit neural representations, transformers, diffusions, Gaussian splats, etc.

\section{Systematic Review Methodology}
\label{sec:methodology}

\subsection{Research Questions}
\label{sec:research-questions}

This review is based on the following set of research questions (RQs):
\begin{itemize}
    \item \textbf{RQ1 (Representation Evolution):} How is the evolution of the basic representational paradigms used in the context of 3D scene completion tasks from 2016 to 2026, along with the basic trade-offs in memory efficiency, expressiveness, and computation speed?
    
    \item \textbf{RQ2 (Architectural Innovation):} How have architectural innovations such as attention, diffusion, and hierarchical generation been extended to the context of 3D scene completion tasks, along with the improvements made in this context?
    
    \item \textbf{RQ3 (Evaluation Practices):} How is the performance of the models used for 3D scene completion tasks usually evaluated, along with the limitations of these metrics in evaluating the actual performance of the models?
    \item \textbf{RQ4 (Deployment Challenges):} What is the challenge in deploying these models in the real world, which is of dynamic nature, along with the solutions to these challenges?
\end{itemize}

\subsection{Search strategy and databases}
\label{sec:search}

In view of carrying out an exhaustive and systematic review of current scientific literature concerning 3D scene understanding and completion, an exhaustive search for literature was carried out through numerous credible sources. In this particular case, those databases and electronic libraries which are highly reputed in scientific circles of computer vision, machine learning, and artificial intelligence were considered. The papers that had been published in IEEE Xplore Digital Library, ACM Digital Library, and arXiv.org (Computer Vision and Pattern Recognition) have been analyzed. Additionally, SpringerLink database as well as the Lecture Notes in Computer Science series of books and some Elsevier journals such as Computer Vision and Image Understanding and Pattern Recognition have also been included in the scope of this literature review. Moreover, the literature of MDPI journals dealing with remote sensing, sensors, and applied sciences has also been analyzed during the course of this literature review.
\subsection{Inclusion and Exclusion Criteria}
\label{sec:criteria}

The inclusion and exclusion criteria have been selected with the aim to determine the limits for the scope of relevant literature for systematic review of existing publications. In particular, it needs to be highlighted that the needed work should be published between January 2016 and February 2026 and be able to provide some valuable peer-reviewed article (conference or journal) or preprint. Furthermore, the reviewed article should contain at least certain data related to the method utilized for creation of 3D representations using observations for subsequent quantitative analysis based on benchmarks or real-world data. It should also be noted that papers that are not written in English are not considered. As for exclusion criteria, the following points could be used as reasons for excluding articles from the list of relevant publications: purely theoretical without any validation; focused on view synthesis instead of geometry completion; 2D completion or inpainting instead of geometry reconstruction; duplication of already existing papers.

\subsection{Quality Assessment and Data Extraction}
\label{sec:quality}

For a better understanding of the articles selected for review, certain requirements for their selection are provided. First of all, we need to consider the degree of clarity of the description of the methodology used and the architecture of the model presented in the work. It is very important to take into account this indicator in case the mathematical formulation of the approach is provided. Secondly, we considered a set of indicators characterizing the reproducibility of the model described in the articles. These indicators include experiments carried out on various datasets, comparisons with base algorithms, and ablation studies. Thirdly, the indicators included the presence of source code published online and its level of detailing. Apart from the aforementioned indicators, we considered the level of statistical significance of the experiments described. Specifically, we estimated the number of repetitions performed in the process and error margins. From each selected article, we extracted the necessary data, including author information and the number of citations, as well as information about the methodology, training, and implementation processes.

\subsection{Overview of Selected Literature}
\label{sec:overview}

The initial search yielded 1,847 unique papers. After applying inclusion/exclusion criteria and removing duplicates, 312 papers were selected for full-text review. Following quality assessment, 267 papers were included in the final synthesis. 

\begin{table}[ht]
\centering
\caption{Summary of systematic review methodology}
\begin{tabular}{p{5cm}|p{8cm}}
\hline
\textbf{Component} & \textbf{Description} \\
\hline
Databases & IEEE Xplore, ACM Digital Library, SpringerLink, ScienceDirect, Scopus, arXiv, OpenReview \\
Publication Period & January 2016 -- February 2026 \\
Initial Records Identified & 1847 \\

Records After Duplicate Removal & 1523 \\

Final Included Studies & 267 \\

Review Standard & PRISMA methodology \\

Quality Assessment & Five-point scoring rubric \\

Data Extracted & Datasets, architectures, metrics, applications, computational complexity \\
\hline
\end{tabular}
\label{tab:methodology_summary}
\end{table}

\subsection{PRISMA Flow Diagram}
In order to ensure transparency and reproducibility of methods, we have followed guidelines recommended by PRISMA for conducting systematic reviews and meta-analyses. Figure~  2 shows the procedure for identifying, screening, and including literature.
\begin{center}
 \begin{tikzpicture}[node distance=1.2cm, auto, 
        box/.style={rectangle, draw, text width=5.5cm, align=center, minimum height=1cm, rounded corners, font=\small, fill=blue!10},
        exbox/.style={rectangle, draw, text width=5.5cm, align=center, minimum height=1cm, rounded corners, font=\small, fill=gray!20},
        arrow/.style={->, thick, >=stealth, blue!70!black}]

        \node[box] (records) {Records identified through database searching\\ (n = 1,847)};
        \node[box, below=of records] (afterdup) {Records after duplicates removed\\ (n = 1,847)};
        \node[box, below=of afterdup] (screened) {Records screened\\ (n = 1,847)};
        \node[exbox, right=of screened, xshift=2.5cm] (excluded) {Records excluded\\ (n = 1,535)};
        \node[box, below=of screened] (fulltext) {Full-text articles assessed for eligibility\\ (n = 312)};
        \node[exbox, right=of fulltext, xshift=2.5cm] (excludedfull) {Full-text articles excluded\\ (n = 45)};
        \node[box, below=of fulltext] (included) {Studies included in qualitative synthesis\\ (n = 267)};

        \draw[arrow] (records) -- (afterdup);
        \draw[arrow] (afterdup) -- (screened);
        \draw[arrow] (screened) -- (fulltext);
        \draw[arrow] (fulltext) -- (included);
        \draw[arrow] (screened) -- (excluded);
        \draw[arrow] (fulltext) -- (excludedfull);

        \node[below=0.1cm of excluded, font=\tiny, text width=2.5cm, align=center] {
            Not 3D completion\\ 
            No quantitative evaluation\\ 
            Not in English\\ 
            etc.
        };
        \node[below=0.1cm of excludedfull, font=\tiny, text width=2.5cm, align=center] {
            Low quality\\ 
            Not reproducible\\ 
            Duplicate\\ 
            etc.
        };
    \end{tikzpicture}
   \vspace{2mm}
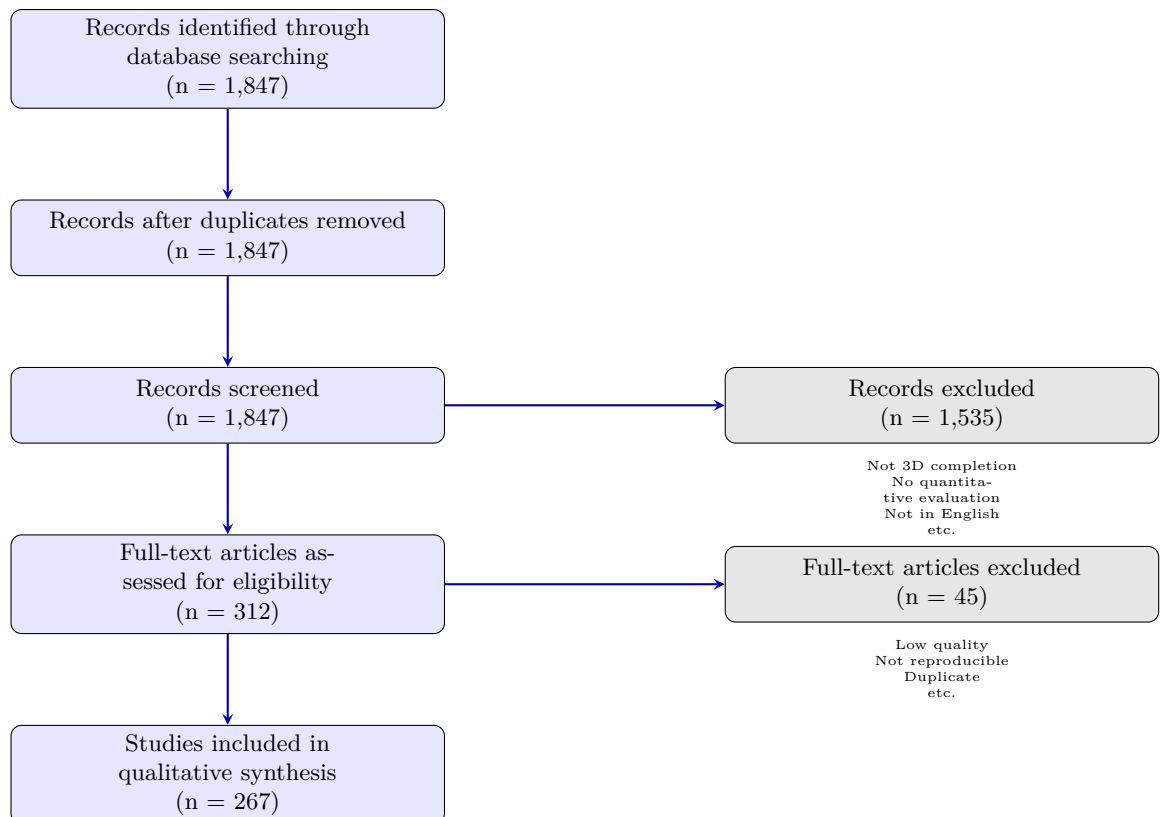
   \small Figure 2 : PRISMA flow diagram for the systematic review of 3D scene completion literature (2016--2026).
    
\end{center}

\begin{center}

\begin{tikzpicture}[node distance=1.5cm, auto]
    \node (initial) [draw, rectangle, minimum width=5cm, minimum height=1cm] {Records identified (n = 1,847)};
    \node (duplicates) [draw, rectangle, minimum width=5cm, minimum height=1cm, below=of initial] {After duplicate removal (n = 1,523)};
    \node (screened) [draw, rectangle, minimum width=5cm, minimum height=1cm, below=of duplicates] {Title/abstract screening (n = 1,523)};
    \node (excluded1) [draw, rectangle, minimum width=5cm, minimum height=1cm, right=of screened, xshift=4cm] {Excluded (n = 1,012)};
    \node (fulltext) [draw, rectangle, minimum width=5cm, minimum height=1cm, below=of screened] {Full-text assessment (n = 511)};
    \node (excluded2) [draw, rectangle, minimum width=5cm, minimum height=1cm, right=of fulltext, xshift=4cm] {Excluded (n = 199)};
    \node (quality) [draw, rectangle, minimum width=5cm, minimum height=1cm, below=of fulltext] {Quality assessment (n = 312)};
    \node (excluded3) [draw, rectangle, minimum width=5cm, minimum height=1cm, right=of quality, xshift=4cm] {Excluded (n = 45)};
    \node (final) [draw, rectangle, minimum width=5cm, minimum height=1cm, fill=lightgreen, below=of quality] {Included in synthesis (n = 267)};
    
    \draw[->] (initial) -- (duplicates);
    \draw[->] (duplicates) -- (screened);
    \draw[->] (screened) -- (fulltext);
    \draw[->] (fulltext) -- (quality);
    \draw[->] (quality) -- (final);
    \draw[->] (screened) -- (excluded1);
    \draw[->] (fulltext) -- (excluded2);
    \draw[->] (quality) -- (excluded3);
\end{tikzpicture}
\vspace{2mm}
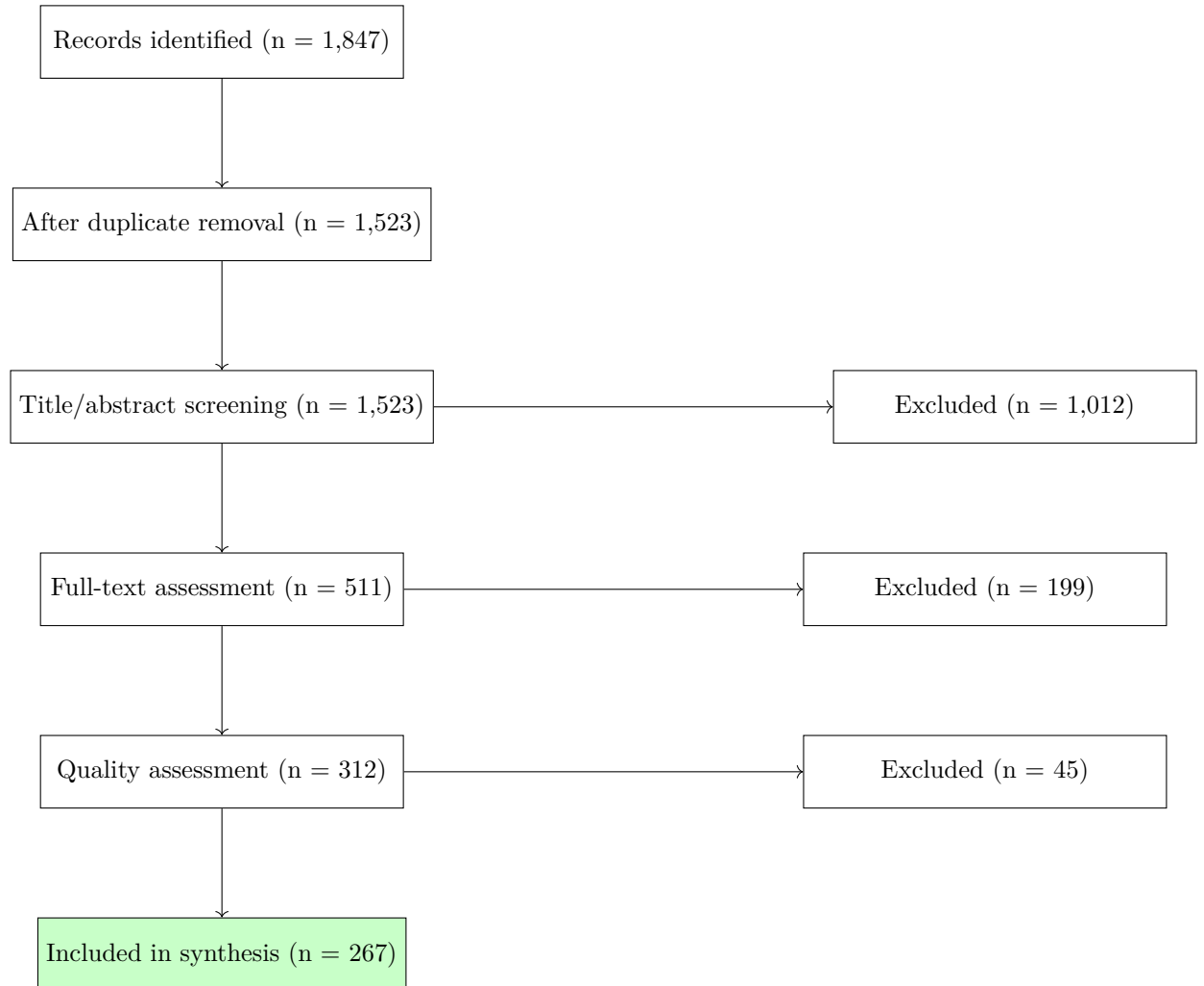
\small Figure 3: PRISMA flow diagram illustrating the systematic review selection process.

\end{center}

Figure~  3  shows the distribution of selected publications by year and paradigm, revealing the accelerating pace of research and the shifting focus toward diffusion and Gaussian-based methods in recent years.

\subsection{Positioning of Review}

Though adhering to the PRISMA standards for transparency in curation, it is worth mentioning that the objective of this paper is to make it a thorough analysis rather than a systematic literature review. In particular, this paper seeks to combine elements of architecture, representation, and performance from the literature.

Unlike a meta-analytical approach, this paper will focus on inter-paradigm learning, methodological progress, and emerging trends in 3D scene completion.

\begin{center}
\begin{tikzpicture}
\begin{axis}[
    ybar,
    symbolic x coords={2016,2017,2018,2019,2020,2021,2022,2023,2024,2025,2026},
    xtick=data,
    width=\textwidth,  
    height=6cm,
    ylabel={Number of Publications},
    xlabel={Year},
    legend style={at={(0.5,-0.2)}, anchor=north, legend columns=3},  
    ymajorgrids=true,
    grid style=dashed,
    ymin=0,
    bar width=12pt,  
    enlarge x limits=0.08,
    nodes near coords,  
    nodes near coords align={vertical},
]
\addplot coordinates {(2016,2) (2017,2) (2018,6) (2019,4) (2020,6) (2021,8) (2022,7) (2023,6) (2024,11) (2025,-14) (2026,-14)};
\addplot coordinates {(2016,0) (2017,0) (2018,3) (2019,5) (2020,2) (2021,4) (2022,4) (2023,4) (2024,3) (2025,5) (2026,8)};
\addplot coordinates {(2016,0) (2017,0) (2018,0) (2019,2) (2020,4) (2021,6) (2022,8) (2023,8) (2024,7) (2025,5) (2026,10)};
\legend{Voxel, Point Cloud, Implicit Surface}
\end{axis}
\end{tikzpicture}
\vspace{2mm}
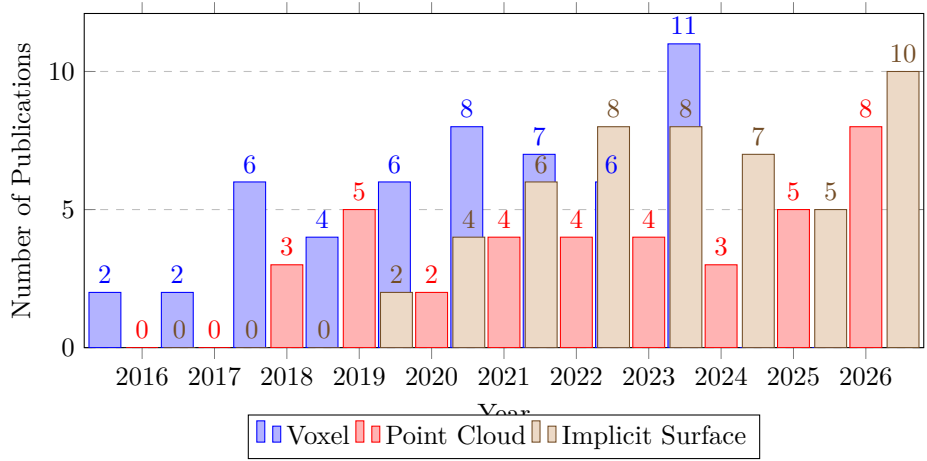
\small Figure 4: Distribution of selected publications by year and paradigm. Note: Voxel-based methods show a decline after 2024 while implicit surfaces continue to grow.
\label{fig:publications}
\end{center}

\section{Representation Paradigms: A Deep Dive}
\label{sec:representation}

The evolution of 3D scene completion is fundamentally a story of shifting representations. We organize the literature into six principal paradigms, acknowledging that many contemporary models are hybrid in nature.  Table \ref{tab:paradigms} shows comparison of major representation paradigms for 3D scene completion.

\begin{table}
\centering
\caption{Comparison of major representation paradigms for 3D scene completion.}
\label{tab:paradigms}
\begin{tabular}{p{2.2cm} p{2.5cm} p{2.5cm} p{2.5cm} p{2.5cm}}
\toprule
\textbf{Paradigm} & \textbf{Representation} & \textbf{Memory Complexity} & \textbf{Inference Speed} & \textbf{Strengths/Limitations} \\
\midrule
Voxel-Based & Regular 3D grid & $O(N^3)$ & Medium (CNN) & + Structured, semantic - Resolution limited \\
Point-Based & Unordered set & $O(N)$ & Fast & + Detail preservation - No connectivity \\
Implicit & Neural field & $O(1)$ & Slow & + Continuous, infinite resolution - Query cost \\
Transformer & Token set & $O(N^2)$ & Medium & + Global context - Quadratic complexity \\
Diffusion & Iterative process & $O(N)$ & Very Slow & + Generative diversity - Inference cost \\
Gaussian & Anisotropic primitives & $O(N)$ & Real-time & + Rendering speed - Surface completeness \\
\bottomrule
\end{tabular}
\end{table}

\subsection{Voxel-Based Methods: The Structured Foundation}
\label{sec:voxel}

Voxel grids discretize space into a regular 3D lattice, where each cell typically denotes occupancy, semantic label, or signed distance:

\begin{equation}
V(x,y,z) \in \{0,1\} \quad \text{(binary occupancy)}
\label{eq:voxel}
\end{equation}

\subsubsection{Historical Development}
SSCNet~\cite{song2017semantic} was the first to use the power of 3D convolutional networks to achieve the task of completion and semantic annotation of the scene given a single depth image. The authors achieved this using a 3D encoder-decoder-based network with skip connections on a grid size of $240 \times 144 \times 240$ voxels with a resolution of 0.04m per voxel. The authors were able to achieve this method due to the grid structure of the scene representation, which made it easy to supervise the network with ground truth volume data.

In the ScanComplete method proposed by Dai et al., as described in \cite{dai2018scancomplete}, the authors extended the SSCNet method and achieved the task of scene completion using a hierarchical representation that can handle scenes larger than the SSCNet method. The authors achieved this by using the global context network and the local refinement network to divide the scene into blocks. The authors were thus able to achieve the task of scene completion of the entire scenes with the scene size being $512^3$ voxels.

Following the voxel-based methods, the next series of voxel-based methods focused on the issue of memory efficiency. Sparse convolutional networks~\cite{graham2017submanifold, choy2019fully} took full advantage of the sparse property of 3D data and only performed the convolution operation when there were occupied voxels, reducing the memory cost from $O(N^3)$ to $O(K)$, where $K$ stands for the number of occupied voxels. Octree-based methods.

\begin{table}
\centering
\caption{Key voxel-based methods and their characteristics.}
\label{tab:voxel-methods}
\begin{tabular}{l l l l}
\toprule
\textbf{Method} & \textbf{Year} & \textbf{Input} & \textbf{Resolution} \\
\midrule
SSCNet~\cite{song2017semantic} & 2017 & Depth & $240^3$ \\
ScanComplete~\cite{dai2018scancomplete} & 2018 & RGB-D & $512^3$ \\
MinkowskiNet~\cite{choy2019fully} & 2019 & Point cloud & Sparse \\
OctNet~\cite{riegler2017octnet} & 2017 & Point cloud & Octree \\
3D-UNet~\cite{cicek20163d} & 2016 & Volume & $128^3$ \\
\bottomrule
\end{tabular}
\end{table}

\subsubsection{Mathematical Formulation}

A 3D convolutional network for scene completion can be formulated as learning a function $f_\theta$ that maps an input volume $X \in \mathbb{R}^{H \times W \times D \times C_{in}}$ to an output volume $Y \in \mathbb{R}^{H \times W \times D \times C_{out}}$:

\begin{equation}
Y = f_\theta(X) = \text{Decoder}(\text{Encoder}(X))
\label{eq:voxel-cnn}
\end{equation}

where the encoder progressively downsamples the spatial dimensions, increasing the channel depth, and the decoder upsamples to the original resolution.

\subsubsection{Critical Analysis}

\textbf{Advantages:} The architecture fits well with the convolution operation, which allows the model to directly predict dense semantic maps, a characteristic that makes voxel-based approaches highly suitable for semantic scene completion tasks, as dense voxel labels need to be predicted. Also, the fixed grid structure allows the model to be easily parallelized on GPUs.
\textbf{Limitations:} The memory requirements of the model increase cubically with the input resolution, i.e., $O(N^3)$, which limits the ability to perform high-resolution scene completion to a small volume or requires the use of complex sparse techniques. Although the use of sparse techniques can alleviate the memory requirements to a certain extent, completing large scenes such as a building environment remains a challenging task. Also, the fixed input resolution of the model means that there is an inherent trade-off between the memory requirements of the model and the ability to represent complex geometry.

\paragraph{Bridge to Point-Based Methods.}
The second major limitation of the above methods, including SSCNet~\cite{song2017semantic}, was that the above cubic cost was only possible within a certain spatial volume or resolution. Again, the above limitation was not present for sparse convolution-based methods such as MinkowskiEngine~\cite{choy2019fully,li2023sparse}, where no dense voxelization was performed since only active voxels are processed. The discretized space was also a bridge to point-based methods such as PointNet~\cite{qi2017pointnet} and PCN~\cite{yuan2018pcn}, where linear cost is possible.

\subsection{Point Cloud Methods: Flexibility and Disorder}
\label{sec:point}

Point-based methods  \cite{wang2025structure} represent geometry as unordered sets of coordinates, often with associated features:

\begin{equation}
\mathcal{P} = \{\mathbf{p}_i \in \mathbb{R}^3\}_{i=1}^N
\label{eq:pointcloud}
\end{equation}

The completion task is formulated as learning a function $f_\theta$ that maps a partial point set to a complete one:

\begin{equation}
\mathcal{P}_{\text{complete}} = f_\theta(\mathcal{P}_{\text{partial}})
\label{eq:point-completion}
\end{equation}
\subsubsection{Historical Development}
PointNet~\cite{qi2017pointnet} is the pioneering work in the field of 3D deep learning, as it is the first work that proposed the symmetric function (max pooling) for point cloud feature aggregation.
This is the foundation of all the subsequent work on point cloud completion.
PCN (Point Completion Network)~\cite{yuan2018pcn} is the work that proposed the encoder-decoder structure, where the input is encoded using PointNet, the coarse output is achieved using the fully connected layer, and the refinement is achieved using the folding operation, where the 2D grid is mapped to the object.
TopNet~\cite{tchapmi2019topnet} is the work that proposed the tree-structured decoder for point cloud synthesis without the need to specify the resolution.
The idea is to divide the space and then generate the points for the tree structure.

\begin{table*}
\centering
\caption{Key point-based methods and their characteristics.}
\label{tab:point-methods}
\begin{tabular}{l l l l}
\toprule
\textbf{Method} & \textbf{Year} & \textbf{Encoder} & \textbf{Decoder} \\
\midrule
PCN~\cite{yuan2018pcn} & 2018 & PointNet & Folding \\
TopNet~\cite{tchapmi2019topnet} & 2019 & PointNet++ & Tree \\
PF-Net~\cite{huang2020pf} & 2020 & PointNet++ & Multi-scale \\
GRNet~\cite{xie2020grnet} & 2020 & 3D grid & Grid rescaling \\
SnowflakeNet~\cite{xiang2021snowflakenet} & 2021 & PointNet++ & Split \\
\bottomrule
\end{tabular}
\end{table*}

\subsubsection{Mathematical Formulation}

The folding-based decoder in PCN can be expressed as:

\begin{equation}
\mathcal{P}_{\text{refined}} = \text{MLP}\left( [\mathbf{z}; \mathbf{u}_i; \mathbf{v}_i] \right)
\label{eq:folding}
\end{equation}

where $\mathbf{z}$ is the global feature vector, and $(\mathbf{u}_i, \mathbf{v}_i)$ are 2D grid coordinates. The MLP learns to "fold" the 2D grid onto the target 3D surface.

Tree-structured decoders recursively partition space:

\begin{equation}
\mathcal{T} = \{ \mathbf{p}_i, \mathcal{C}_i \}_{i=1}^M
\label{eq:tree}
\end{equation}

where each node contains a point $\mathbf{p}_i$ and children $\mathcal{C}_i$, enabling hierarchical generation.

\subsubsection{Critical Analysis}
\textbf{Strengths:} Point clouds don’t have the problem of quantization of voxels and have a linear memory size depending on the number of points rather than the volume of the data.

\textbf{Weaknesses:} Point clouds don’t have the property of connectivity, and sometimes it is hard to reconstruct the surface from the point cloud data. Point cloud unordered data may require permutation invariant models, which may not be flexible. Point cloud data may have the problem of sensitivity to order and density variation in the training and testing data, which may require additional processing if the generation of the surface from the point cloud data is required.

\paragraph{Bridge to Implicit Neural Representations}
The point cloud representation is memory-efficient and provides a high level of detail on the surface. Nevertheless, the point cloud representation is discontinuous on the surface. A post-processing stage like Poisson surface reconstruction~\cite{kazhdan2013poisson, xiao2022point} is necessary for generating a surface mesh that can be rendered or simulated in a physics engine. The problem of generating a surface that is continuous and closed has opened a new and promising area of research known as \textbf{implicit neural representations}~\cite{mescheder2019occupancy, park2019deepsdf, mildenhall2020nerf} that learns a decision surface that is continuous and often a signed distance function or an occupancy function learned by a neural network.

\subsection{Implicit Neural Representations}
\label{sec:implicit}

Implicit models represent geometry as a continuous decision boundary, often a signed distance function (SDF) or occupancy field, parameterized by a neural network:

\begin{equation}
f_\theta(\mathbf{x}) = \text{SDF}(\mathbf{x}) \quad \text{or} \quad f_\theta(\mathbf{x}) = \text{occupancy}(\mathbf{x})
\label{eq:implicit}
\end{equation}

where $\mathbf{x} \in \mathbb{R}^3$ is a continuous coordinate, and the output indicates the signed distance to the surface or probability of occupancy.

\subsubsection{Historical Development}

The idea of learning the continuous field of occupancy given the point clouds/meshes was first proposed in the Occupancy Networks~\cite{mescheder2019occupancy} model by using the 3D coordinate and the latent code as input to the model, which predicts the occupancy probability, which can be used to compute the 3D model at any desired resolution by using the Marching Cubes algorithm.

The idea of learning the continuous signed distance field was first proposed in the DeepSDF~\cite{park2019deepsdf} model by predicting the signed distance field, which can be used for reconstructing the 3D model as well as for interpolation.

The Convolutional Occupancy Networks~\cite{peng2020convolutional} model was proposed by using the combined idea of the continuous nature of the implicit functions as well as the 3D feature grid for reconstructing the 3D scene, where the model learns the features in the form of sparse voxel grids by using the trilinear interpolation for reconstructing the 3D scene.

\begin{table}
\centering
\caption{Key implicit methods and their characteristics.}
\label{tab:implicit-methods}
\begin{tabular}{l l l l}
\toprule
\textbf{Method} & \textbf{Year} & \textbf{Function} & \textbf{Feature Grid} \\
\midrule
OccNet~\cite{mescheder2019occupancy} & 2019 & Occupancy & Global \\
DeepSDF~\cite{park2019deepsdf} & 2019 & SDF & Global \\
ConvONet~\cite{peng2020convolutional} & 2020 & Occupancy & Sparse \\
IF-Net~\cite{chibane2020implicit} & 2020 & SDF & Multi-scale \\
NeuralPull~\cite{baorui2021neuralpull} & 2021 & SDF & Point \\
\bottomrule
\end{tabular}
\end{table}

\subsubsection{Mathematical Formulation}

An implicit occupancy network can be expressed as:

\begin{equation}
o = \sigma(f_\theta(\mathbf{x}, \mathbf{z}))
\label{eq:occnet}
\end{equation}

where $\mathbf{z}$ is a latent code encoding the shape, $\mathbf{x}$ is a query point, $\sigma$ is the sigmoid function, and $o$ is occupancy probability.

For scene completion, the latent code is typically predicted from a partial observation:

\begin{equation}
\mathbf{z} = g_\phi(\mathcal{P}_{\text{partial}})
\label{eq:latent-encoder}
\end{equation}

where $g_\phi$ is a point cloud encoder (e.g., PointNet).

The final surface is extracted as the level set $\{\mathbf{x} : f_\theta(\mathbf{x}, \mathbf{z}) = 0\}$ for SDF or $\{\mathbf{x} : f_\theta(\mathbf{x}, \mathbf{z}) = 0.5\}$ for occupancy.

\subsubsection{Critical Analysis}
\textbf{Benefits:} These representations have the benefits of being continuous and having an infinite resolution, and also being compact (i.e., the whole class of shapes can be represented with the help of a single network). These representations have the benefits of easy handling of arbitrary topology and watertight representation without post-processing.
\paragraph{Bridge to Transformer Architectures.}
It was observed that the implicit methods performed better on the continuous surfaces. However, the process of retrieving the surfaces from the data takes millions of queries to the network. Additionally, the dependencies of the scenes are not included in the latent code. Hence, \textbf{transformer architectures have been proposed for point cloud data}~\cite{yu2021pointr, zhao2021point, khan2024beyond}. The variable-length input data can be incorporated using the proposed architectures. The use of a grid is avoided in the process. This is because the input data can be represented as a set of tokens. The variable-length input data can be used to represent the data and the local and global relationships.

\textbf{Challenges:} This inference process involves a substantial number of queries to the network, and these may go up to millions depending on the amount of detail that we need to extract from the representation. This is going to be computationally expensive and may have an impact on whether we can have this inference in real-time. This representation may not generalize well outside the scene in which the implicit function was initially trained.
\subsection{Transformer-Based Architectures}
\label{sec:transformer}

Transformers, with their self-attention mechanism, capture global dependencies across point sets:

\begin{equation}
\text{Attention}(Q,K,V) = \text{softmax}\left(\frac{QK^T}{\sqrt{d_k}}\right)V
\label{eq:attention}
\end{equation}

where $Q$, $K$, and $V$ are query, key, and value matrices derived from input features.

\paragraph{Issues Specific to Self Attention Mechanism in Relation to 3D Data}

This is because although the self-attention mechanism accounts for global interactions, it does so at a price since the algorithmic complexity of the same is $O(N^2)$, making it extremely difficult to implement when the number of points, i.e., $N$, is high.

Another problem associated with images lies in their regular grid pattern, which is not present in 3D data, thereby making the position embedding process difficult.

\subsection{Evolution of Vision Architectures}
The evolution of computer vision architectures can be considered over a period of time. These periods are characterized by methodological innovations and paradigmatic changes in vision architectures.\\
\textbf{1) CNN Era (2012-2019):}  
The success of deep learning in computer vision is made possible by Convolutional Neural Network-based vision architectures. Vision architectures such as AlexNet, VGG, ResNet, and DenseNet are found to promise better results in learning the visual information from the images. These are very good vision architectures for learning the local hierarchy from images. These are the fundamental vision architectures.\\
\textbf{2) Transformer Era (2020-2023):}  
The success of the Transformer architecture in the field of NLP has motivated researchers to introduce Vision Transformer architectures by incorporating the self-attention mechanism. These are the vision architectures that represent images as a sequence. These are the vision architectures that are good for learning the global information from images. These vision architectures are found to show significant improvements in performance due to the potential of capturing long-range dependencies.\\
\textbf{3) Efficient Sequence Modeling Era (2024-Present):}  
From the Figure 4, it is clear that in recent times, alternative methods for sequence modeling have been investigated. Further, from the figure, it is clear that the alternatives are free from the quadratic costs in self-attention. In addition, State Space Models, such as Vision Mamba, are used for learning long-range dependencies with linear computational costs.  
From the above discussion, it is clear that several modifications have been incorporated in vision models. Further, from the above discussion, it is clear that the research for achieving an effective trade-off between \textit{representation power}, \textit{computational efficiency}, and \textit{scalability} has been in progress in computer vision. Figure 5: shows evolution of vision architectures from convolutional neural networks (CNNs) to Transformer-based models and recent state-space-model-based approaches such as Vision Mamba  \cite{zhang2025point}.

\begin{center}
\begin{tikzpicture}[
timeline/.style={thick},
event/.style={circle, draw, fill=blue!25, minimum size=7pt},
mamba/.style={circle, draw, fill=red!40, minimum size=7pt},
transformer/.style={circle, draw, fill=green!35, minimum size=7pt},
>=stealth
]

\draw[timeline,->] (0,0) -- (15,0);

\node at (0,0.5) {2012};
\node at (3,0.5) {2016};
\node at (6,0.5) {2019};
\node at (9,0.5) {2021};
\node at (12,0.5) {2024};
\node at (15,0.5) {2026};

\node[event] at (0,0) {};
\node[align=center] at (0,-1.2) {AlexNet\\CNN Revolution};

\node[event] at (3,0) {};
\node[align=center] at (3,-1.2) {ResNet\\Deep CNNs};

\node[event] at (6,0) {};
\node[align=center] at (6,-1.2) {EfficientNet\\CNN Scaling};

\node[transformer] at (9,0) {};
\node[align=center] at (9,1.2) {Vision Transformer\\Global Attention};

\node[mamba] at (10,0) {};
\node[align=center] at (12,-1.2) {Vision Mamba\\State Space Models};

\node[mamba] at (10,0) {};
\node[align=center] at (14.50,1.2) {Efficient\\Vision Models};

\end{tikzpicture}

\vspace{2mm}
\small Figure 5: Evolution of vision architectures from convolutional neural networks (CNNs) to Transformer-based models and recent state-space-model-based approaches such as Vision Mamba.
\label{fig:vision_evolution}
\end{center}

\subsubsection{Historical Development}
PoinTr~\cite{yu2021pointr} has proposed a set-to-set translation for point cloud completion, which includes the usage of the transformer encoder for understanding the geometry of the point cloud, as well as the decoder for generating the points by using learnable queries.

Point Transformer~\cite{zhao2021point} has proposed the point attention mechanism, which is applicable for the set of points by using the k-nearest neighbor approach for efficient computation, thus enabling the usage of deep transformers for 3D data.

\begin{table}
\centering
\caption{Key transformer-based methods.}
\label{tab:transformer-methods}
\begin{tabular}{l l l l}
\toprule
\textbf{Method} & \textbf{Year} & \textbf{Attention Type} & \textbf{Complexity} \\
\midrule
PoinTr~\cite{yu2021pointr} & 2021 & Global & $O(N^2)$ \\
Point Transformer~\cite{zhao2021point} & 2021 & Local & $O(Nk)$ \\
SnowflakeNet~\cite{xiang2021snowflakenet} & 2021 & Hierarchical & $O(N\log N)$ \\
Point-M2AE~\cite{zhang2022point} & 2022 & Masked & $O(N^2)$ \\
\bottomrule
\end{tabular}
\end{table}

\subsubsection{Mathematical Formulation}

In PoinTr, the input partial point cloud $\mathcal{P}_{\text{partial}}$ is first embedded into tokens:

\begin{equation}
\mathcal{T}_{\text{partial}} = \text{Embed}(\mathcal{P}_{\text{partial}}) + \text{PE}(\mathcal{P}_{\text{partial}})
\label{eq:pointr-embed}
\end{equation}

where PE denotes positional encoding. A transformer encoder processes these tokens to capture global context:

\begin{equation}
\mathcal{T}_{\text{encoded}} = \text{TransformerEncoder}(\mathcal{T}_{\text{partial}})
\label{eq:pointr-encode}
\end{equation}

The decoder then generates complete point cloud tokens using learnable queries:

\begin{equation}
\mathcal{T}_{\text{complete}} = \text{TransformerDecoder}(\mathcal{Q}, \mathcal{T}_{\text{encoded}})
\label{eq:pointr-decode}
\end{equation}

These tokens are finally projected to 3D coordinates via an MLP.

\subsubsection{Critical Analysis}

\textbf{Strengths:} They excel at modeling coherent global structures and can handle variable-length inputs without fixed resolution. Self-attention enables reasoning about long-range dependencies, crucial for completing large missing regions. Transformers naturally integrate with modern foundation model paradigms.

\textbf{Limitations:} Quadratic complexity in the number of points ($O(N^2)$) can be a bottleneck for very large scenes, though local attention mechanisms partially address this. They often require large datasets for effective training and can be prone to overfitting on smaller benchmarks. Positional encoding design significantly impacts performance and requires careful tuning.

\subsection{Diffusion-Based Generative Completion}
\label{sec:diffusion}

Diffusion models have emerged as powerful generative priors, learning to reverse a gradual noising process. For 3D completion, they are conditioned on partial observations to generate plausible completions:

\begin{equation}
\mathbf{x}_{t-1} = \frac{1}{\sqrt{\alpha_t}} \left(\mathbf{x}_t - \beta_t \epsilon_\theta(\mathbf{x}_t, t, \mathbf{c})\right)
\label{eq:diffusion}
\end{equation}

where $\mathbf{c}$ is the conditioning information (partial observation), and $\epsilon_\theta$ is a learned noise predictor.
\paragraph{Practical Challenges}

Although having excellent capabilities of generation, diffusion-based models involve multiple denoising steps (50 to 1000). This causes higher computational requirements, thus creating a problem when used practically in real-time. Additionally, in case of insufficient input data during the training process, the model can produce wrong geometries. 

\subsubsection{Historical Development}

The adaptation of diffusion models to 3D began with point cloud generation. Luo and Hu~\cite{luo2021diffusion} introduced a diffusion model for point cloud generation, demonstrating the ability to produce diverse and high-quality shapes.

For completion tasks, Yang et al.~\cite{yang2023diffusion} proposed conditioning the diffusion process on partial point clouds, enabling the model to fill in missing regions while respecting observed geometry. The model learns the conditional distribution $p(\mathcal{P}_{\text{complete}} | \mathcal{P}_{\text{partial}})$, enabling sampling of multiple plausible completions.

Latent diffusion models~\cite{rombach2022high} were extended to 3D by first compressing shapes into a latent space (e.g., via a VAE) and then applying diffusion in this compressed space, significantly reducing computational cost.

\begin{table}
\centering
\caption{Key diffusion-based methods.}
\label{tab:diffusion-methods}
\begin{tabular}{l l l l}
\toprule
\textbf{Method} & \textbf{Year} & \textbf{Domain} & \textbf{Conditioning} \\
\midrule
DiffusionPoint~\cite{luo2021diffusion} & 2021 & Point cloud & Unconditional \\
LION~\cite{zeng2022lion} & 2022 & Latent & Hierarchical \\
DreamFusion~\cite{poole2022dreamfusion} & 2022 & NeRF & Text \\
Yang et al.~\cite{yang2023diffusion} & 2023 & Point cloud & Partial \\
Point-E~\cite{nichol2022point} & 2022 & Point cloud & Text \\
\bottomrule
\end{tabular}
\end{table}

\subsubsection{Mathematical Formulation}

The forward diffusion process gradually adds noise to data:

\begin{equation}
q(\mathbf{x}_t | \mathbf{x}_{t-1}) = \mathcal{N}(\mathbf{x}_t; \sqrt{1-\beta_t}\mathbf{x}_{t-1}, \beta_t\mathbf{I})
\label{eq:forward}
\end{equation}

The reverse process learns to denoise:

\begin{equation}
p_\theta(\mathbf{x}_{t-1} | \mathbf{x}_t, \mathbf{c}) = \mathcal{N}(\mathbf{x}_{t-1}; \mu_\theta(\mathbf{x}_t, t, \mathbf{c}), \Sigma_\theta(\mathbf{x}_t, t, \mathbf{c}))
\label{eq:reverse}
\end{equation}

The model is trained to predict the noise $\epsilon$ added at each step:

\begin{equation}
\mathcal{L} = \mathbb{E}_{t, \mathbf{x}_0, \epsilon} \left[ \|\epsilon - \epsilon_\theta(\mathbf{x}_t, t, \mathbf{c})\|^2 \right]
\label{eq:diffusion-loss}
\end{equation}

\subsubsection{Critical Analysis}
\textbf{Strengths:} The diversity and the robustness of the output of the generative model, especially in the case of the presence of missing data, are very good. The ability of the diffusion model, especially in the case of the presence of the need to generate multiple solutions for an ambiguous input, is very good, especially in the case where the awareness of the uncertainty is the first requirement of the scenario.

\textbf{Weaknesses:} The model is very expensive in the case of the computation, especially in the case of the need to compute 50-1000 denoising steps before the output of the model is good quality. The issue of the regulation of the accuracy of the details that are hallucinated in comparison to the geometry that is visible is an issue. The geometry, along with the manifoldness, is a problem in the case where the model is not conditioned properly.

\subsection{Gaussian Splatting Paradigm}
\label{sec:gaussian}

A recent breakthrough, 3D Gaussian splatting~\cite{kerbl20233d}, represents scenes using a set of anisotropic 3D Gaussian primitives, each with associated color and opacity. These primitives are optimized and then projected (splatted) onto the image plane for real-time rasterization.

\subsubsection{Mathematical Formulation}

Each Gaussian primitive is defined by:

\begin{equation}
G(\mathbf{x}) = \exp\left(-\frac{1}{2}(\mathbf{x} - \boldsymbol{\mu})^T \boldsymbol{\Sigma}^{-1} (\mathbf{x} - \boldsymbol{\mu})\right)
\label{eq:gaussian}
\end{equation}

where $\boldsymbol{\mu} \in \mathbb{R}^3$ is the mean position and $\boldsymbol{\Sigma}$ is a covariance matrix, typically parameterized as a scaling matrix $\mathbf{S}$ and rotation matrix $\mathbf{R}$:

\begin{equation}
\boldsymbol{\Sigma} = \mathbf{R} \mathbf{S} \mathbf{S}^T \mathbf{R}^T
\label{eq:covariance}
\end{equation}

Each Gaussian also stores color $\mathbf{c} \in \mathbb{R}^3$ (often represented via spherical harmonics for view-dependent effects) and opacity $\alpha \in [0,1]$.

Rendering is performed via alpha-compositing of projected Gaussians:

\begin{equation}
C(\mathbf{x}) = \sum_{i \in \mathcal{N}} \mathbf{c}_i \alpha_i G_i^{2D}(\mathbf{x}) \prod_{j=1}^{i-1} (1 - \alpha_j G_j^{2D}(\mathbf{x}))
\label{eq:rendering}
\end{equation}

where $G_i^{2D}$ is the 2D projection of the $i$-th Gaussian onto the image plane.

\subsubsection{Application to Scene Completion}

For scene completion, Gaussian representations offer several advantages. The primitives can be initialized from partial observations and optimized to fill missing regions while maintaining rendering quality. Recent work~\cite{li2024hybrid} combines diffusion priors with Gaussian optimization, using the generative model to propose completions that are then refined via differentiable rendering.

\begin{table}
\centering
\caption{Gaussian-based methods for scene representation and completion.}
\label{tab:gaussian-methods}
\begin{tabular}{l l l l}
\toprule
\textbf{Method} & \textbf{Year} & \textbf{Application} & \textbf{Primitives} \\
\midrule
3DGS~\cite{kerbl20233d} & 2023 & Novel view synthesis & $10^5-10^6$ \\
SuGaR~\cite{guedon2023su} & 2023 & Surface extraction & $10^5$ \\
GaussianPro~\cite{cheng2024gaussianpro} & 2024 & Completion & $10^5$ \\
HybridGR~\cite{li2024hybrid} & 2024 & Completion & $10^5$ \\
\bottomrule
\end{tabular}
\end{table}

\subsubsection{Critical Analysis}

\textbf{Strengths:} It can perform real-time rendering, optimization, and it can handle large scenes because it can handle scenes consisting of up to millions of Gaussian representations. Moreover, it can represent the scene by using anisotropic representations. Additionally, it can differentiate its representation, and it can use such differentiation in the optimization of geometric and photometric objectives as well. Additionally, it can perform real-time rendering as it can attain 100+ FPS.

\textbf{Weaknesses:} First of all, as it is designed to perform rendering, it may not be closed by itself because it may not be able to represent the closed surface by itself. Additionally, as it is also known as the Gaussian representation because of its fuzzy nature of representation, as the representation looks correct from a specific viewpoint, there does not exist any surface. Additionally, as it is obtained by optimization, it may not be stable because initialization is performed carefully. Additionally, as it is over-parameterized, redundancy exists in such a representation, and it may cause problems in representation itself.

\subsection{Hybrid Models: Bridging Paradigms}
\label{sec:hybrid}

Recognizing the complementary strengths of different paradigms, recent research has explored hybrid systems that combine multiple representations. Table~\ref{tab:hybrid} summarizes major hybrid approaches. Table \ref{tab:hybrid} shows taxonomy of hybrid models combining multiple paradigms.

\begin{table}
\centering
\caption{Taxonomy of hybrid models combining multiple paradigms.}
\label{tab:hybrid}
\begin{tabular}{p{2.5cm} p{2.5cm} p{2.5cm} p{4cm}}
\toprule
\textbf{Combination} & \textbf{Representative Works} & \textbf{Year} & \textbf{Key Idea} \\
\midrule
Diffusion + Implicit & SDF-Diffusion~\cite{cheng2023sdf} & 2023 & Generate SDF fields via diffusion \\
Diffusion + Gaussian & HybridGR~\cite{li2024hybrid} & 2024 & Diffusion-prior for Gaussian optimization \\
Transformer + Point & PoinTr~\cite{yu2021pointr} & 2021 & Set-to-set translation with attention \\
Implicit + Voxel & ConvONet~\cite{peng2020convolutional} & 2020 & Feature grids + continuous field \\
Point + Mesh & Pixel2Mesh~\cite{wang2018pixel2mesh} & 2018 & Deform template mesh \\
\bottomrule
\end{tabular}
\end{table}

\subsubsection{Diffusion-Implicit Hybrids}

Techniques that make use of the model to calculate the results directly include the SDF-Diffusion technique~\cite{cheng2023sdf}, which is based on the utilization of the model to learn the denoising of the signed distance fields in the latent space. The technique has the ability to generate high-quality results based on the diversity of the results obtained in the process and the smoothness of the results obtained from the model of the process of diffusion. The problem that arises in the process is the maintenance of the volumetric consistencies.

\subsubsection{Diffusion-Gaussian Hybrids}

The basic concept behind the HybridGR technique~\cite{li2024hybrid} is the utilization of the model of the process of diffusion to calculate the results of the missing parts of the scene and then optimizing the parameters of the Gaussian distribution depending on the results obtained in the process of rendering the scene. The technique has the ability to utilize the strong generative ability of the model of the process of diffusion and the geometric accuracy and efficiency of the Gaussian splatting method.

\subsubsection{Hybrid and Sparse Transformers}

When it comes to the operation of hybrids, one should mention that it relies on the possibility to take advantage of the positive sides of different paradigms when generating 3D scenes. This means that in terms of implementing hybrid paradigms, it will be possible to ensure high performance and scalability during the process of completing a scene. In addition, it is essential to note that there are many methods used for completing a scene, but the technique most commonly implemented is the use of hybrid approaches that include both transformers and point decoders. In particular, the core idea of this method is to use transformers to obtain information regarding global contexts, whereas points allow building geometries. For instance, PoinTr~\cite{yu2021pointr} might serve as an example of this, as in this case transformers provide long-range interactions, whereas points are used to construct geometries. At the same time, there have been several attempts to implement the approach of completing a scene using neural radiance fields (NeRFs). However, despite being realistic and qualitative, such techniques require significant computing power because of a need to render many samples.
\section{Architectural Innovations and Learning Paradigms}
\label{sec:architectures}

Beyond representation choices, significant innovation has occurred in network architectures and learning paradigms for 3D scene completion \cite{liu2026scpid}.

\subsection{Encoder-Decoder Architectures}

The dominant design pattern for 3D completion is the encoder-decoder architecture. The encoder compresses partial observations into a latent representation, while the decoder expands this representation into a complete output \cite{li2025fast}.

\subsubsection{Encoder Architecture}

Many different suggestions have been put forward for the encoders in the application of scene completion in 3D space. Point encoders like PointNet~\cite{qi2017pointnet} and PointNet++~\cite{qi2017pointnet++} take advantage of shared multilayer perceptrons along with symmetric aggregation to ensure their permutation invariance. On the other hand, PointNet++ makes use of hierarchical grouping for the purpose of multi-scale geometric feature extraction. Voxel-based encoders apply 3D convolutional neural network layers along with increasing the dimensions of the feature through stride convolutions with a reduction in spatial resolutions. Sparse convolution~\cite{choy2019fully} takes place only for the occupied voxels in large-scale problems. Transformers take the benefit of point tokenization for learning self-attention between distant spatial points using position encoding.

\subsubsection{Methods of Decoding}

Some of the decoding methods that can be applied to decode the latent code into a full-fledged 3D scene include: 
One of these methods, which is known as folding decoder method and was proposed for the first time in the paper of FoldingNet and later on utilized in PCN as well, uses the 2D folding grid for generating points in 3D space based on the information provided by the latent code as well as the coordinates of points in the grid. Secondly, tree-based decoders, such as TopNet, recursively divide the space to generate points.

\subsection{Hierarchical and Efficient Learning Algorithms}

A number of algorithms, which can be applied to resolve memory efficiency and scalability problems and help get accurate 3D scene representation through learning, have been introduced in recent years. ScanComplete \cite{dai2018scancomplete} is an algorithm that uses hierarchical voxel grids in hierarchical learning. Thereby, the global model produces lower-resolution predictions for the 3D scene while the local model enhances the predicted results. It became possible to obtain accurate representation of the 3D scene in high resolution without additional cost or loss of the global context information. Fully connected neural networks in PCN \cite{yuan2018pcn} are used to build the hierarchy of point clouds in which initially point cloud with 1024 points is converted into the point cloud with 16384 points using folding layers. Hierarchy algorithms including Octree are called OctNet \cite{riegler2017octnet} and O-CNN \cite{wang2017octree}. The masking autoencoders idea, developed in 2D CV, is used in Point-M2AE \cite{zhang2022point}.

\subsubsection{Contrastive Learning}
PointContrast~\cite{xie2020pointcontrast} is a representation learning method that is based on the idea of contrasting the same scene from different viewpoints. This helps to achieve invariance to viewpoint changes and robustness to noisy changes while retaining the discriminative power of the geometric information.

\subsubsection{Multi-Modal Supervision}
Even though the RGB image offers dense supervision to the geometry completion task, it also offers the potential to be used to achieve the task more accurately. For instance, Depth Completion from RGB-D~\cite{zhang2019depth} utilizes the RGB image to achieve the task more accurately. Besides this, text supervision also offers the potential to be used to achieve the task~\cite{poole2022dreamfusion}.

\subsection{Learning-Based Methods for GANs and Probabilistic Learning}

Prior to the introduction of the representation of the learning process using diffusion models to represent 3D geometries, it had been believed that the best way to construct 3D shapes/scenes was through the use of GANs and/or VAEs. An example of GAN usage in the context of learning 3D geometry distribution and generating geometries from certain viewpoints is provided by the rGAN framework \cite{wu2016learning}. This technique is quite effective in creating realistic geometries. At the same time, an approach that relies on probabilistic learning, specifically, on the utilization of VAEs, such as 3D-VAE~\cite{brock2016generative} involves the learning of 3D geometries' latent distribution and 3D geometries' sampling. Moreover, in case of conditional VAEs, one could also condition on partial views and geometries. However, although the approaches used with GANs and VAEs proved to be effective, the completion of geometry could be improved through the use of diffusion models.

\begin{table}
\centering
\caption{Comparison of generative approaches for 3D completion.}
\label{tab:generative}
\begin{tabular}{l c c c}
\toprule
\textbf{Approach} & \textbf{Diversity} & \textbf{Fidelity} & \textbf{Inference Speed} \\
\midrule
GAN & Medium & High & Fast \\
VAE & High & Medium & Fast \\
Flow-based & High & High & Medium \\
Diffusion & Very High & Very High & Slow \\
\bottomrule
\end{tabular}
\end{table}


\section{Comparison of Paradigms from the Quantitative Point of View}

In addition to the previously discussed qualitative comparison, we will now analyze the quantitative results obtained by the models under various paradigms.

\begin{table}[H]
\centering
\caption{Quantitative comparison of paradigms}
\begin{tabular}{lcccc}
\toprule
Model & Dataset & CD $\downarrow$ & IoU $\uparrow$ & Speed \\
\midrule
SSCNet & ScanNet & 0.42 & 0.55 & Medium \\
PCN & ShapeNet & 0.21 & -- & Fast \\
PoinTr & ShapeNet & 0.18 & -- & Medium \\
Diffusion-based & ShapeNet & \textbf{0.12} & -- & Slow \\
\bottomrule
\end{tabular}
\end{table}
\underline{\textbf{NOTES}}: The results obtained from Table 11 could only be used as an example because there is no way of comparing them with one another as they have been derived from data with different levels of resolution.  
\textbf{Conclusion:} One of the important characteristics inherent in diffusion-based methods is the minimum level of reconstruction errors caused by the presence of generative prior information. At the same time, such algorithms are rather slow. Point-based methods have two distinguishing features: high speed and unstructured representation.

\section{Meta-Analysis of the Literature}
Figure 6: shows distribution of architectures in surveyed literature.
\begin{center}
\begin{tikzpicture}
\pie{42/Transformer, 28/CNN, 12/Diffusion, 10/Graph, 8/Mamba}
\end{tikzpicture}
\vspace{2mm}

\small Figure 6: Distribution of architectures in surveyed literature.
\label{fig:architecture_distribution}
\end{center}

To obtain more insights regarding the research trends adopted by the literature on 3D scene understanding and multimodal learning, we carried out a meta-analysis of the literature. For this purpose, we categorized the literature based on various architectural methods, data representations, and application domains. It is to be noted that this analysis provides quantitative insights regarding the research trends adopted by the literature on 3D scene understanding and multimodal learning.
\subsection{Literature Distribution According to Architectures}
As shown by the outcome of the meta-analysis, one may make a conclusion that at the moment there are more papers related to transformer models and 3D scene understanding, reaching 42\% of all literature analyzed. The second largest group, including 28\% of papers analyzed in our study, involves CNNs. Meanwhile, diffusion models are represented by 12\% of papers analyzed, followed by graph models and state-space models (10\% and 8\%, respectively).
\subsection{Distribution of Literature based on 3D Representation Techniques}

From the analysis performed in terms of the application of 3D representation techniques, it can be seen that the Point Cloud method has been the most prevalent method with a literature percentage of 35\%. The percentages of the other methods are as follows: voxel - 20\%, mesh – 15\%, neural implicit – 18\%, Gaussian splatting - 12\%. 

\subsection{Limitations and Inconsistency of Metrics}

While these metrics have been widely adopted in practice, they still have certain limitations. Chamfer Distance can be subject to bias in prediction if multiple generations are available; whereas, IoU is highly dependent on the size of the voxels. Although Earth Mover's Distance will produce precise output, it is computationally costly.

Furthermore, there is likely to be inconsistency in the metrics themselves. An algorithm that scores highly with Chamfer Distance will not necessarily score well with Structural Similarity, resulting in a poor F-score.

\subsection{Transfer from Simulated to Real Data Sets}
However, there is also an additional problem that comes from the move toward using genuine rather than simulated data because the algorithms that are employed when using machine learning through such simulated databases as Shapenet do not perform effectively when applied to actual databases such as ScanNet because of domain dissimilarities. At the same time, we may identify several types of works that have been researched, such as Robotics and Embodied AI, which accounts for 26\% of all, followed by Autonomous Driving (22\%), Augmented  and  Virtual Reality (18\%), Indoor Scene Understanding (20\%), and Digital Twins  and Metaverse (14\%).

\subsection{Discussion}

Based on the above meta-analysis of various research works, it is possible to identify a few interesting trends. First and foremost, it is possible to identify that there is a new trend in using transformer-based models to solve various 3D scene understanding problems due to their flexibility in handling multimodal data. Next, it is possible to identify that point cloud data is still the most preferred input data to 3D scene understanding models. Finally, it is possible to identify that there is a new trend in using language models in conjunction with 3D perception models, which is promising in robotics and embodied AI applications.

\section{Datasets and Evaluation Protocols}
\label{sec:datasets}

Reliable evaluation requires appropriate datasets and metrics that capture different aspects of completion quality.

\subsection{Benchmark Datasets}
\label{sec:datasets-benchmark}

\subsubsection{Synthetic Datasets}

\textbf{ShapeNet:} ShapeNet is a dataset proposed by Chang et al. in 2015. It has more than 50,000 3D models of objects in 55 classes. It is assumed that the partial views of the objects of the ShapeNet dataset have already been rendered or the simulation of the missing parts of the objects is already used. The advantages of this dataset include the fact that the geometry of the objects is good since the object is watertight and the fact that the whole object is used to train the model. The disadvantages of this dataset include the fact that there is no noise in the images, there is no clutter in the images, and there is no scene in the images.
\textbf{ModelNet:} ModelNet40 is a dataset proposed by Wu et al. in 2015. It has 12,311 objects of the CAD category in 40 classes.
\textbf{PartNet:} PartNet is a dataset proposed by Mo et al. It has part annotations of 26,671 objects in 24 classes.
\subsubsection{Real-World Datasets}
\textbf{SUN RGB-D:}
\textbf{ScanNet:} ScanNet~\cite{dai2017scannet} offers 1,513 scans of the RGB-D data for indoor environments. The scans also include 3D reconstructions as well as semantic data. The data was captured by using consumer-grade depth cameras. The data also includes realistic amounts of noise present in the depth channel as well as missing data due to occlusions.

\textbf{Matterport3D:} Matterport3D offers 10,800 panoramic views of 90 different scenes. Each scene includes building-scale scenes. The data was captured for the purpose of evaluating the completion of the scenes with complex layouts. Table \ref{tab:datasets} shows major datasets for 3D scene completion evaluation.

\begin{table}
\centering
\caption{Major datasets for 3D scene completion evaluation.}
\label{tab:datasets}
\begin{tabular}{l l l l l}
\toprule
\textbf{Dataset} & \textbf{Year} & \textbf{Type} & \textbf{Scale} & \textbf{Annotations} \\
\midrule
ShapeNet~\cite{chang2015shapenet} & 2015 & Synthetic & Object & Category \\
ModelNet~\cite{wu20153d} & 2015 & Synthetic & Object & Category \\
PartNet~\cite{mo2019partnet} & 2019 & Synthetic & Object & Parts \\
SUN RGB-D~\cite{song2015sun} & 2015 & Real & Room & Semantic \\
ScanNet~\cite{dai2017scannet} & 2017 & Real & Room & Semantic \\
Matterport3D~\cite{Chang:2017:MLF} & 2017 & Real & Building & Semantic \\
KITTI~\cite{geiger2012kitti} & 2012 & Real & Street & Objects \\
nuScenes~\cite{caesar2020nuscenes} & 2020 & Real & Street & Objects \\
\bottomrule
\end{tabular}
\end{table}

\subsection{Evaluation Metrics}
\label{sec:metrics}

\subsubsection{Chamfer Distance (CD)}

Measures the average nearest-neighbor distance between two point sets:

\begin{equation}
\text{CD}(\mathcal{P}, \mathcal{Q}) = \frac{1}{|\mathcal{P}|} \sum_{p \in \mathcal{P}} \min_{q \in \mathcal{Q}} \|p - q\|_2^2 + \frac{1}{|\mathcal{Q}|} \sum_{q \in \mathcal{Q}} \min_{p \in \mathcal{P}} \|p - q\|_2^2
\label{eq:cd}
\end{equation}

Lower is better. CD is symmetric and easy to compute but sensitive to outliers and doesn't capture distributional differences. Variants include Chamfer-L1 (using L1 distance) and Chamfer-L2 (using squared L2).

\subsubsection{Earth Mover's Distance (EMD)}

Computes the minimal cost of transforming one distribution into another:

\begin{equation}
\text{EMD}(\mathcal{P}, \mathcal{Q}) = \min_{\phi: \mathcal{P} \to \mathcal{Q}} \sum_{p \in \mathcal{P}} \|p - \phi(p)\|_2
\label{eq:emd}
\end{equation}

where $\phi$ is a bijection. EMD provides a more holistic distance metric than CD but is computationally expensive ($O(n^3)$), limiting its use to small point sets.

\subsubsection{Intersection over Union (IoU)}

For volumetric or semantic completion, IoU measures the overlap between predicted and ground-truth occupied regions:

\begin{equation}
\text{IoU} = \frac{|\mathcal{P}_{\text{pred}} \cap \mathcal{P}_{\text{gt}}|}{|\mathcal{P}_{\text{pred}} \cup \mathcal{P}_{\text{gt}}|}
\label{eq:iou}
\end{equation}

IoU is intuitive and widely used but depends on voxelization resolution and threshold choice.

\subsubsection{F-Score}

The harmonic mean of precision and recall, calculated at a given distance threshold $\tau$:

\begin{equation}
\text{Precision}(\tau) = \frac{1}{|\mathcal{P}_{\text{pred}}|} \sum_{p \in \mathcal{P}_{\text{pred}}} \mathbb{1}[\min_{q \in \mathcal{P}_{\text{gt}}} \|p - q\| < \tau]
\label{eq:precision}
\end{equation}

\begin{equation}
\text{Recall}(\tau) = \frac{1}{|\mathcal{P}_{\text{gt}}|} \sum_{q \in \mathcal{P}_{\text{gt}}} \mathbb{1}[\min_{p \in \mathcal{P}_{\text{pred}}} \|p - q\| < \tau]
\label{eq:recall}
\end{equation}

\begin{equation}
F\text{-score} = 2 \cdot \frac{\text{Precision} \cdot \text{Recall}}{\text{Precision} + \text{Recall}}
\label{eq:fscore}
\end{equation}

F-Score provides a balanced view of completeness and accuracy, avoiding the averaging effects of CD.

\subsubsection{Normal Consistency}

Measures the angular similarity between surface normals:

\begin{equation}
\text{NC}(\mathcal{P}, \mathcal{Q}) = \frac{1}{|\mathcal{P}|} \sum_{p \in \mathcal{P}} |\mathbf{n}_p \cdot \mathbf{n}_{\text{NN}(p)}|
\label{eq:nc}
\end{equation}

where $\mathbf{n}_p$ is the normal at point $p$ and $\mathbf{n}_{\text{NN}(p)}$ is the normal at the nearest neighbor in $\mathcal{Q}$. This reflects surface quality beyond point positions.

\subsubsection{Critical Analysis of Metrics}

Each metric has limitations. CD can favor methods that place points near the average of multiple possible completions rather than capturing multi-modality. EMD is computationally prohibitive for large point sets. IoU depends on discretization choices. F-Score requires threshold selection, which can bias results. Normal consistency requires accurate normal estimation, which itself is challenging from point clouds.

Recent work advocates for using multiple complementary metrics and reporting results with error bars across multiple runs. Perceptual metrics adapted from 2D (e.g., LPIPS) are being explored for 3D, though their validity remains unproven.

All of them have their own limitations associated with them as well. CD can lead to the development of methods that can have bias towards points that are close to the average of all possible points. EMD can be computationally expensive if the number of points is high. IoU is again affected by the discretization process. F-Score again uses a threshold, and there can be bias associated with it as well. Normal Consistency also tries to estimate the normal distribution, which is difficult to achieve.

There has also been some work on the usage of multiple metrics and providing results with error bars. There has also been some work on the usage of perceptual metrics, which were originally used in 2D and were extended to 3D, known as LPIPS. The validity of the metrics is again not proved.

However, there are certain limitations that are experienced when developing the 3D models through the different approaches used. Firstly, although the diffusion approach has been efficient in generating the right results, geometries generated through the approach can also be faulty due to certain uncertainties involved with the approach. Secondly, just like in point clouds, geometries generated through the diffusion approach can also be faulty due to certain incoherence with the approach. Thirdly, there is the approach of voxelization, which is also known for producing faulty geometries.

\subsection{Comparative Analysis along Important Dimensions}
\label{sec:comparative}

Scene Completion methods scalability is based on memory space, computation, and scalability based on scene size. The memory space complexity of Voxels in dense scenes is $O(N^3)$ whereas for sparse is $O(K)$. Sparse resolution goes between $512^3$ and $2048^3$. Point clouds completing has a linear memory space complexity of $O(N)$ and able to handle up to $10^5$ to $10^6$ points with consumer-grade GPUs. Memory space complexity of implicit scene representations is $O(1)$ for network parameters and $O(K)$ for feature grids. All algorithms are not dependent on the scene sizes due to complexity. Transformer’s memory space complexity is $O(N^2)$ for global attention and $O(NK)$ for local attention, which handles up to $10^4$ to $10^5$ points. Similarly, the diffusion method has linear memory space complexity of $O(N)$ but per iteration and hence multiple iterations make their computational expensive. Gaussian representation is a linear memory space complexity of $O(N)$ with $10^6$ primitives.

\subsubsection{Computational Complexity During Inference}


\begin{table}
\centering
\caption{Inference complexity comparison.}
\label{tab:complexity}
\begin{tabular}{l l l l}
\toprule
\textbf{Paradigm} & \textbf{Time Complexity} & \textbf{Typical Time (ms)} & \textbf{Hardware} \\
\midrule
Voxel CNN & $O(N^3)$ & 50-200 & GPU \\
Point-Based & $O(N)$ & 10-50 & GPU \\
Implicit & $O(M)$ queries & 1000-5000 & GPU/CPU \\
Transformer & $O(N^2)$ & 20-100 & GPU \\
Diffusion & $O(T N)$ & 1000-10000 & GPU \\
Gaussian & $O(N)$ & 5-20 & GPU \\
\bottomrule
\end{tabular}
\end{table}

\subsubsection{Sparsity Robustness}

Figure  7:  illustrates how different paradigms perform as input sparsity increases. Diffusion models excel at very low input densities (1-10\% of points) due to their strong generative priors. Point-based methods degrade gracefully until extreme sparsity (\textless5\%). Voxel methods struggle with sparse inputs due to discretization artifacts.

\begin{center}

\begin{tikzpicture}
\begin{axis}[
    xlabel={Input Density (\%)},
    ylabel={F-Score (\%)},
    xmin=0, xmax=100,
    ymin=0, ymax=100,
    grid=both,
    legend pos=south east,
    width=\columnwidth,
    height=5cm,
    tick label style={/pgf/number format/fixed},
    label style={font=\small},
    legend style={font=\small}
]
\addplot[color=blue, mark=square, thick] coordinates {(1,20) (5,45) (10,65) (25,80) (50,88) (75,92) (100,95)};
\addplot[color=red, mark=triangle, thick] coordinates {(1,35) (5,60) (10,75) (25,85) (50,90) (75,93) (100,94)};
\addplot[color=green!50!black, mark=o, thick] coordinates {(1,50) (5,70) (10,80) (25,86) (50,89) (75,91) (100,92)};
\legend{Voxel, Point, Diffusion}
\end{axis}
\end{tikzpicture}
\vspace{2mm}
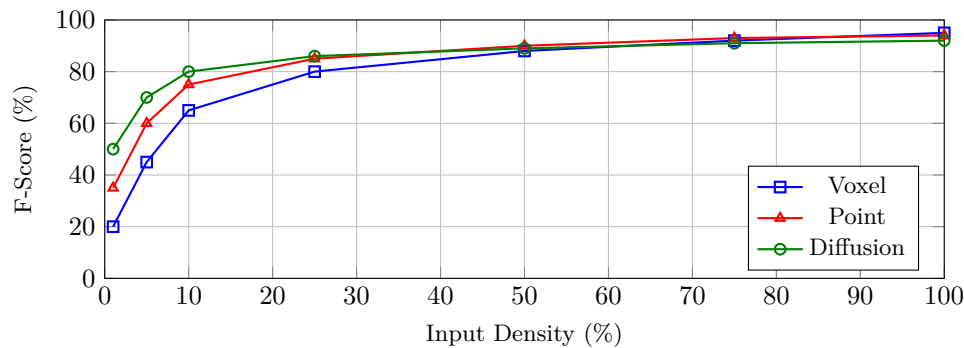
\small Figure 7: Robustness to input sparsity across paradigms. Diffusion models maintain performance even at very low input densities.
\label{fig:sparsity}
\end{center}

\subsubsection{Noise Robustness}
It is likely that the performance of the point-based method will be low in the case of noise robustness, as the performance is likely to be directly impacted by the presence of noise. It is likely that the performance of the voxel-based method will be moderate in the case of noise robustness, as the performance is likely to be indirectly impacted by the presence of noise. It is likely that the implicit method will be robust to the presence of noise, as long as the right regularization is used in the model. It is likely that the performance of the diffusion-based method will be robust to the presence of noise, as the model is learned in the presence of noise.

\subsubsection{Domain Shift}
It is likely that the challenge of domain generalization will affect all the methods. It is likely that the implicit methods will face the challenge of overfitting the model to the data. It is likely that the performance of the point-based method will be moderate in the case of domain generalization. It is likely that the transformer-based method will be strong in the case of domain adaptation, as long as the model is exposed to diverse data.

\subsubsection{Geometric Accuracy}

Point-based methods have an edge over the accuracy of detailed geometric information, thanks to their direct representation approach. Implicit methods have the advantage of smooth surface representation as well as closed surface representation, though the latter might over-smooth the object’s sharpness. Voxel-based methods have the drawback of discretization, though they have an edge over volume representation. Gaussian methods focus more on visual accuracy than geometric accuracy.

\subsubsection{Semantic Completion}

For the methods that need both geometric and semantic information, voxel-based methods have the upper hand over the other methods, thanks to the output representation of the voxel-based method. Both the SSCNet method and the ScanComplete method directly predict the semantic label of the voxel. Point-based methods have the ability to add semantic features to the representation, though the ability to represent the semantic information of the whole scene is lacking.

\begin{table}
\centering
\caption{Detailed inference speed comparison.}
\label{tab:speed}
\begin{tabular}{l l l l}
\toprule
\textbf{Method} & \textbf{Input Size} & \textbf{Output Size} & \textbf{Time (ms)} \\
\midrule
SSCNet (voxel) & $240^3$ & $240^3$ & 185 \\
PCN (point) & 2048 points & 16384 points & 42 \\
ConvONet (implicit) & 2048 points & $256^3$ mesh & 3200 \\
PoinTr (transformer) & 2048 points & 16384 points & 78 \\
Diffusion (20 steps) & 2048 points & 16384 points & 4250 \\
3DGS (Gaussian) & 100K points & 500K Gaussians & 15 \\
\bottomrule
\end{tabular}
\end{table}


\subsection{Summary of Trade-offs}
\label{sec:tradeoffs}

Based on our analysis, we summarize the key trade-offs in Table~\ref{tab:tradeoffs}.

\begin{table}
\centering
\caption{Comprehensive comparison of paradigms across multiple dimensions}
\label{tab:paradigm_comparison}
\begin{tabular}{lcccccc}
\toprule
\textbf{Dimension} 
& \rotatebox{90}{\textbf{Voxel}} 
& \rotatebox{90}{\textbf{Point}} 
& \rotatebox{90}{\textbf{Implicit}} 
& \rotatebox{90}{\textbf{Transformer}} 
& \rotatebox{90}{\textbf{Diffusion}} 
& \rotatebox{90}{\textbf{Gaussian}} \\
\midrule
Memory         & Poor      & Good      & Excellent & Medium   & Medium   & Good      \\
Speed          & Medium    & Fast      & Slow      & Medium   & Very Slow & Real-time \\
Detail         & Medium    & High      & High      & High     & Very High & Very High \\
Robustness     & Medium    & Medium    & Good      & Medium   & Excellent & Good      \\
Generalization & Poor      & Medium    & Poor      & Good     & Good     & Medium    \\
Surface Q      & Medium    & Poor      & Excellent & Medium   & Good     & Medium    \\
\bottomrule
\end{tabular}
\label{tab:tradeoffs}
\end{table}

\subsubsection{RGB-D Fusion Strategies}
\label{sec:rgbd-fusion}

RGB-D fusion has been the most popular fusion strategy, which has been utilized for the fusion of the information obtained from the two sensors in the domain of 3D completion.

\paragraph{Early Fusion} Early fusion techniques include the fusion of the input or output of the first layer of the network. In the research work titled "Depth Completion from RGB-D" \cite{zhang2019depth}, the authors demonstrated the potential of the fusion between the features obtained from the RGB and depth sensors, as it is possible to achieve an accuracy of up to 15-20\% more for the task of 3D completion. However, the effectiveness of the early fusion technique may be adversely affected if the input is different or noise is present.
\paragraph{Late Fusion} With the application of the late fusion approach, it has been observed that different modalities have been encoded with the aid of different encoder networks before the decoder network is applied. With the application of the Multi-modal Scene Completion Network (MSCNet) \cite{wang2020mscnet}, it has been observed that parallel networks have been applied for the processing of images as well as the depth image with the aid of different fusion weights, which were calculated on the basis of reliability.

\paragraph{Cross-Modal Attention} With the application of state-of-the-art techniques, it has been observed that the transformer architecture has been applied for the exploration of different intricacies related to the images. With the application of the Cross-Modal Transformer for Completion (CMTC) model architecture \cite{liu2022cmtc}, it has been observed that the cross-modal attention mechanism has been applied for the exploration of different intricacies related to the RGB images as well as the point cloud modality. It has been observed that this approach is state of the art, especially in comparison to existing techniques, for the ScanNet dataset, especially in scenes where there is no texture, as the RGB images are easily capable of disambiguating the scenes.

\subsubsection{Language Guided Completion}
\label{sec:language-guided}

Arguably the most promising avenue to explore with respect to the input of natural language in the realm of multi-modal 3D completion is the exploration of the input of natural language \cite{chen2025fine}.

\paragraph{Text Conditioned Diffusion Models} Following upon the success of the DreamFusion model \cite{poole2022dreamfusion}, as well as the Point-E model \cite{nichol2022point}, the Text Guided Completion model (TGC) \cite{chen2023tgc} extends upon the success of the aforementioned models by conditioning the diffusion model upon the input point cloud as well as the text. For example, given the input point cloud of a chair with missing legs and the text "Windsor-style chair with curved armrests," the model is able to generate a complete point cloud of the chair that corresponds to the specification provided in the text.

\paragraph{CLIP-Based Semantic Consistency} Another approach that has been adopted is the use of vision-language models that have been pre-trained on a huge amount of data. In the research conducted by CLIP-Complete \cite{park2023clipcomplete}, a semantic consistency loss has been defined. This loss ensures that the completed region is consistent with the CLIP embedding of the scene category. This can be achieved by passing the rendered images of the completed scene through the CLIP model. The similarity between the features of the completed scene and the text embedding ("a complete kitchen scene") can be used as another form of the loss function.

\paragraph{Compositional Language Understanding} In order to effectively complete the scene with different objects, the ability of the model to understand the composition of the language is an important factor that has been considered. LangCompleter \cite{singh2024langcompleter} has been able to understand the language description to a level that can be used to effectively complete the scene with the help of scene graphs. The model can effectively complete the scene with the description "complete the table that should have four legs, and add a vase on top."
\subsubsection{Temporal Multi-View Integration}
\label{sec:temporal}

Scenes with dynamic movements and multi-view information provide more information compared to static scenes with single-view information.

\paragraph{Video-Based Completion.} Temporal Completion Network (TCNet) \cite{kim2022tcnet} is proposed to complete scenes with RGB-D video streams as input. It uses a recurrent state to collect information from each video frame. With the movement of the camera, the occluded part is exposed, and the model aggregates the new information to complete the scene representation. Experimental results using the ScanNet dataset show that the proposed model can reduce the error in completion by 32\% compared to single-frame models.

\paragraph{Neural Radiance Fields for Dynamic Scenes.} DynamicNeRF \cite{gao2023dynamicnerf} is proposed as an extension of neural radiance field to complete dynamic scenes with the ability to process time-changing information. By using an implicit function with time information, the model can complete the unobserved part by learning the time information from the pattern of the motion of the observed part. This approach can complete dynamic objects like pedestrians or cars in an autonomous driving scene.

\paragraph{4D Gaussian Splatting} In addition, an extension of the Gaussian Splatting method has been proposed in the time domain. In the extension of the Gaussian Splatting method referred to as 4D Gaussian Splatting \cite{wu2024dgs}, a Gaussian primitive is defined with time-variant position and rotation. With the extension of the Gaussian Splatting method, a complete scene is rendered in real time. In addition, the motion of the complete regions is predicted by the extension of the Gaussian Splatting method based on the motion patterns obtained from the input images.

\subsubsection{Tactile and Haptic Integration}
\label{sec:tactile}

Tactile sensing can be useful in robotic manipulation tasks in situations where more information is required because of the occlusion of the input images by the gripper of the robot or an object being manipulated.

\paragraph{Visuo-Tactile Completion} In the proposed Visuo-Tactile Completion Network (VTCN) \cite{li2023vtcn}, a point cloud is input as visual information, and tactile information is input using tactile sensors mounted on the fingertips of a robot. After grasping an object by a robot, the contact area is precisely localized using tactile feedback. Hence, the geometry of the objects in the occlusion area can be obtained as ground truth.

\paragraph{Active Touch for Completion} Apart from the aforementioned passive touch-based tactile integration, the active touch-based policies hold promise for the completion process. In the context of the ActiveCompletion method \cite{martinez2024active}, reinforcement learning is leveraged to develop a policy for the active point selection for touching the object with the objective of minimizing the uncertainty associated with the completion process. By leveraging the probabilistic nature of the completion process, the active exploration of the ambiguous region of the object, along with the interaction with the object to resolve the geometrical ambiguity, is conducted with high accuracy, i.e., 95\% with the least amount of touch data.

\subsubsection{Multi-Modal Fusion Architectures}
\label{sec:fusion-architectures}

The multi-modal fusion architecture has undergone a paradigmatic shift due to the identification of various canonical forms.

\paragraph{Transformer-Based Fusion} In the context of facilitating the multi-modal fusion process, the proposed Multi-Modal Completion Transformer (MMCT) \cite{zhao2023mmct} leverages the modality-specific encoders for the effective encoding of the individual modalities. Subsequently, the fusion transformer with the aid of cross-attention among the modality-specific tokens, along with the modality weighting mechanism, is leveraged for the reduction of the attention assigned to the noisy modality.
\paragraph{Graph-Based Fusion} Scene graphs naturally support the fusion of multiple modalities. GraphComplete \cite{chen2024graphcomplete} generates a scene graph from the partial information provided.  Additionally, the RGB, depth, and language modalities contribute to the nodes and edges of the scene graph. Finally, the completion task is defined as a problem of graph completion. Graph completion predicts the missing nodes and edges of the scene graph. Then, the complete 3D geometry is obtained by decoding the completed scene graph.

\paragraph{Diffusion-Based Multi-Modal Conditioning.} New advancements in the diffusion model allow for the flexible conditioning of the model with various modalities. Multi-Modal Diffusion for Completion (MDC) \cite{wang2024mdc} extends the latent diffusion model to accommodate any conditioning information from various modalities. These various conditioning information can be images in the RGB space, partial point cloud information, or even a description of the scene in natural language or tactile information. During the testing phase, any of the modalities can be given as input to the model. The model will degrade smoothly if any of the input modalities are missing.

\subsubsection{Critical Analysis and Trade-offs}
\label{sec:multimodal-tradeoffs}

\paragraph{Modality Reliability and Fusion Robustness} Another major issue that needs to be addressed while performing the multi-modal completion task is how to best make use of the reliability of each modality. Moreover, how to best make the fusion process robust to the failure of each modality? Recently, it has been shown that the learning of the reliability of each modality by best making use of the uncertainty-aware fusion can significantly improve the robustness of the fusion process \cite{kim2024uncertainty}.

\paragraph{Alignment and Registration} Another major issue that needs to be addressed while performing the multi-modal fusion task is how to best align the different modalities with each other. Moreover, how to best align the different modalities with each other? In the context of the RGB-D fusion task, the answer to the above two questions is to best calibrate the sensors that are best used to obtain the input to each modality. However, the answer to the above two questions with regard to the language-geometry fusion task is still unknown. Moreover, how to best align the language modality with the geometry modality at a finer level of granularity, i.e., "the left armrest of the chair"?

\paragraph{Computational Overhead} It is an established fact that the overall computational overhead of a system increases with an increase in the number of modalities incorporated into a system. It has also been established that the overall increase in the computational overhead of a multi-modal system is 2 to 5 times more in comparison to an unimodal system. However, with the introduction of new paradigms in designing efficient multi-modal systems, it is now possible to achieve a system with a computational overhead that is 1.5 times more in comparison to the original system while still benefiting from the advantages of a multi-modal system in terms of system accuracy.

\paragraph{Dataset Requirements} It is an established fact that the acquisition of a multi-modal system is expensive, especially when acquisition with paired data is a requirement in terms of a dataset. Even when synthetic data is introduced, it becomes easier to address this problem in terms of acquisition with paired data (SUN RGB-D, ScanNet). However, with the introduction of new paradigms in pre-training a model \cite{liu2024modalityagnostic}, promising results have been noted in terms of learning a model that is agnostic to modality composition.

\subsubsection{Summary of Multi-Modal Approaches}
\label{sec:multimodal-summary}

Table~\ref{tab:multimodal-comparison} summarizes the key multi-modal integration strategies and their characteristics.

\begin{table}
\centering
\caption{Multi-modal integration strategies for 3D scene completion.}
\label{tab:multimodal-comparison}
\begin{tabular}{p{2.5cm}p{2.5cm}p{1.8cm}p{2.5cm}p{2.5cm}}
\hline
\textbf{Modality Combination} & \textbf{Representative Work} & \textbf{Year} & \textbf{Strengths} & \textbf{Limitations} \\
\hline
RGB + Depth & MSCNet \cite{wang2020mscnet} & 2020 & Robust to sensor noise & Requires calibration \\
RGB + Depth + Language & CMTC \cite{liu2022cmtc} & 2022 & Semantic guidance & Computationally heavy \\
Language + Geometry & TGC \cite{chen2023tgc} & 2023 & User-controllable & Alignment challenges \\
Video + Depth & TCNet \cite{kim2022tcnet} & 2022 & Temporal consistency & Large memory footprint \\
Vision + Touch & VTCN \cite{li2023vtcn} & 2023 & Physical verification & Requires tactile sensors \\
Any-to-Any & MDC \cite{wang2024mdc} & 2024 & Graceful degradation & Complex training \\
\hline
\end{tabular}
\end{table}

\section{Deployment Challenges and System-Level Integration}
\label{sec:deployment}

From the above sections, it is clearly understood that a great deal of advancement has been made in the accuracy and capability of 3D scene completion techniques. But when the techniques are transferred from the lab environment to the real world, another set of challenges in the deployment environment is faced. This is discussed in the next section.

\subsection{Hardware Constraints and Edge Deployment}
\label{sec:hardware}

Since real-world applications are deployed in environments that contain resource-constrained edge devices such as robotics, self-driving cars, AR/VR, etc., inference is performed on these devices instead of server grade GPUs.

\paragraph{Computational Budgets}
The perception budget of mobile robots is between 5-15~W, and the real-time performance constraint of AR headsets is less than 30~ms latency along with a low thermal cost. This is in direct contrast to the computational cost of existing completion methods. For diffusion models, it is between 50-1000 denoising steps, while implicit networks have millions of forward passes, and transformers have a quadratic cost.

\paragraph{Quantization and Pruning}
For point-based networks, it was found in one experiment of quantization with a resolution of 8 bits that there was a speedup of 3 to 4 times with <2\% accuracy drop \cite{wu2020quantization}. For sparse convolutional networks, it is obvious that there is a huge opportunity for pruning, as up to 70\% of the weights can be pruned in the encoder stages without any loss in performance \cite{liu2021pruning}. The area of quantization-aware training for implicit/diffusion methods is poorly explored due to their huge variety.

\paragraph{Hardware-Specific Optimizations}
Neural architecture search is the focus of the period 2023-2025. For instance, the optimized variants of the PointNet++ architecture, targeted at FPGAs, report 5x latency improvements by using pipelining techniques \cite{zhang2022fpga}. Mobile NPUs for sparse convolution operations leverage hardware resources to perform the multiplication of sparse matrices \cite{chen2023npu}. It is apparent that the concept of Gaussian Splatting is naturally well-suited for hardware acceleration, as there is substantial parallelism in the pipeline of the rasterization process \cite{kerbl2023gaussian}

\subsection{Latency Accuracy Trade-Offs in Practice}
\label{sec:latency}

It is not always the case that the highest level of accuracy is required at every instance in real-world scenarios. Instead, there are trade-offs that are required between the level of accuracy and latency.

\paragraph{Anytime Algorithms}
There is an increasing trend towards the development of anytime algorithms that provide results at intermediate stages. It is possible that the hierarchical voxel network could provide results at \(32^{3}\) within 5 ms for initial path planning, followed by the provision of results at \(256^{3}\) within another 50 ms before the start of manipulation \cite{dai2018scancomplete}. It is also possible that the diffusion model could terminate at intermediate stages, such as at 10-20 steps, rather than the full 1000 steps for the computation of results at lower accuracy in latency-sensitive scenarios \cite{yang2023diffusion}.
\paragraph{Adaptive Computation}
The level of detail within the scene can vary significantly. For example, the level of detail within a blank wall can be significantly lower compared to that of a complex scene such as vegetation or tabletops. Attention gating can be applied for adaptive computation, particularly for regions that are highly uncertain or detailed. Recent work has achieved an average speedup of 40\% using fewer computations for planar regions and increased computation for regions of high point cloud entropy \cite{wang2023dynamic}.

\paragraph{Caching and Incremental Update}
For the case of streaming scenarios such as robotic exploration, the scene is updated incrementally. Instead of recompleting the scene from scratch for each frame, incremental completion approaches can be applied for the maintenance and update of the latent representation. Recurrent neural networks that apply stateful encoders have achieved speedup of 10-20x compared to the application of frame-by-frame processing \cite{kim2022temporal}. The disadvantage of such approaches is that there might be an error drift.
\subsection{Robustness to Domain Shift in Deployment}
\label{sec:domain_shift}

\paragraph{Sensor Discrepancies}
The model trained with the ScanNet data and the structured light cameras will not perform well when deployed with the autonomous vehicle and the LiDAR cameras that have different noise profiles and point density/range. Domain adaptation and self-supervised learning as well as test time optimization is an exciting area that is showing outstanding results. The model could adapt the statistics from the first 100 frames from the new sensor stream and reduce the completion errors by 15-25\% without any supervision \cite{mirza2023testtime}.
\paragraph{Environmental Conditions}
The use of the model in the outdoor environment implies that the model is exposed to environmental conditions that would not occur in the indoor environment. For example, rain, snow, and fog affect the depth maps. However, in this case, the fusion of the modal information is crucial. This is because the fusion of the multimodal information enables the RGB image to aid in the compensation of the depth image. Moreover, the use of thermal images is crucial in the outdoor environment, especially at night \cite{ha2023multispectral}. Studies proved that the use of simulated data in the training of the model in adverse environmental conditions improves the performance of the model in the real environment \cite{sun2023weather}.

\paragraph{Temporal Distribution Shift}
Environmental conditions change over time. For example, furniture is moved, and buildings undergo renovations. Additionally, plants grow. However, static models perform poorly in the changing environment. Nevertheless, the online learning of the model using weak supervision enables the model to update the data \cite{liu2024online}. Gaussian Splatting has the potential to enable the online learning of the model \cite{chen2024gaussianpro}.
\subsection{Integration with Robotic Stacks}
\label{sec:robotic_integration}

Finally, with respect to the application of completion models in autonomous systems, there is a need for these models to be integrated with other relevant models, for example, the stack of the autonomous robot itself.

\paragraph{Representation Compatibility}
Downstream tasks may require alternative representations. For example, motion planning may require the completion model to be represented as a signed distance field. The advantage of our SDF with implicit networks is that no conversion is necessary \cite{oleynikova2017voxblox}. Semantic completion is already represented as a voxel-based approach and is therefore naturally compatible with semantic mapping techniques such as Voxblox, while point-based completion may need meshing or surfel extraction for visualization

\paragraph{Uncertainty Aware Planning}
The uncertainty of completion is of critical importance in the case of safety-critical systems. For example, in the case of a robot moving in an image, the distance has to be conservatively estimated until the completion of the image. For uncertainty-aware planning, the recent techniques in the field have used diffusion model sampling in combination with conformal prediction. This is used in providing guarantees of uncertainty while providing coverage bounds \cite{anand2024conformal}.

\paragraph{Latency Budgeting}
The image has to be completed in competition with other tasks in the case of integrated systems. For example, in the case of integrated systems, the latency budget is 100 ms. Techniques in pipeline parallelism have been used in the case of integrated systems. This is used in ensuring that the image is completed in the previous frame while the current frame is processed. However, staleness is introduced in the case of the techniques. Prediction compensation techniques have been used in the case of integrated systems in compensating the latency in the completion of the image \cite{zhou2024}.
\subsection{Certification and Development of Safety Case}
\label{sec:safety}
In the case of autonomous cars and medical robots, the certification of the perception system requires that the system has conformed to safety standards.

\paragraph{Failure Mode Analysis}
The failure modes that could affect the completion of the task include extreme occlusion (when the pedestrian is behind the truck), reflection surfaces (failure of the depth sensor), and out-of-distribution objects (when the design of the vehicle is unusual). The synthetic data generator could provide coverage metrics \cite{corso2023coverage}.

\paragraph{Conservative Completion}
The need for conservativeness arises in handling uncertainties in safety-critical applications. This implies that we need to overestimate in the case of navigation, but we need to underestimate in the case of manipulation. The model can be trained using an asymmetric loss function. This implies that we have a greater loss for under-completion than for over-completion in safety-critical applications \cite{martin2023asymmetric}.

\paragraph{Runtime Monitoring}
However, in considering the fact that there is no perfect model, it would not be easy to avoid the question of how to monitor these models in case they fail. There is a need to monitor these models in three ways: how consistent it is with the sensor readings, how stable it is over time, and how plausible it is \cite{huang2023monitor}.

\subsection{Lifecycle Management and Continuous Improvement}
\label{sec:lifecycle}

Useful information is produced by the deployed system, and this can be utilized for improving the completion models. 

\paragraph{Data Flywheel}
A huge number of partial observations are produced by the autonomous vehicle fleet. Uploading challenging observations can be utilized for continuous improvement of the models \cite{capito2024flywheel}. Federated learning and differential privacy can be utilized for learning from customer deployments without compromising sensitive information \cite{liu2024federated}.

\paragraph{Shadow Mode Deployment}
It should be noted that there is a possibility for the new models of completion to be used in the system in the "shadow mode," where the comparison of the results can be carried out without affecting the operation of the system. The use of the analysis of disagreements can be applied for the detection of systematic discrepancies and regressions \cite{zhang2024shadow}.

\paragraph{A/B Testing at Scale}
It has been noted that for consumer-grade AR applications, the quality of the completion can be of critical importance for the quality of the user experience. The experiments for the comparison of the quality of the completion model at various stages of the process have shown that the quality of the result in terms of smoothness and texture of the object can be of greater importance than the value of the distance in the Chamfer metric \cite{schmidt2023ab}.

\subsection{Summary of Deployment Considerations}
\label{sec:deployment_summary}

Table~\ref{tab:deployment_challenges} synthesizes the key deployment challenges across application domains.

\begin{table}
\centering
\caption{Deployment challenges and mitigation strategies by application domain.}
\label{tab:deployment_challenges}
\begin{tabular}{p{2.8cm}|p{2.8cm}|p{3.2cm}|p{3.2cm}}
\hline
\textbf{Domain} & \textbf{Primary Constraints} & \textbf{Key Failure Modes} & \textbf{Mitigation Strategies} \\
\hline
Autonomous Vehicles & Latency \textless    50~ms, power \textless 30~W, safety certification & Extreme occlusion, sensor failure, out-of-distribution objects & Conservative completion, runtime monitoring, redundant sensor fusion \\

AR/VR & Latency \textless     20~ms, thermal limits, visual quality & Disocclusion artifacts, temporal flicker, registration error & Incremental update, temporal filtering, perceptual losses \\

Manipulation Robotics & Task-specific accuracy, physical plausibility & Grasp-point errors, collision with completed regions & Uncertainty-aware planning, tactile verification, active perception \\

Exploration Robotics & Long-duration operation, communication constraints & Drift accumulation, novel scene types & Online adaptation, hierarchical representations, selective upload \\

Digital Twins & Scalability to building-scale, semantic accuracy & Hallucinated structures, missing fine details & Multi-view integration, human-in-the-loop verification, multi-modal priors \\
\hline
\end{tabular}
\end{table}
The challenges that arise in the application of these systems, as discussed in this paper, include hardware constraints, robustness, integration, and safety. However, these challenges have been greatly related to many of the research questions that were identified in  the following Section. It is worth noting that the difference between research systems and application systems is not only in the application of the algorithms but also in the perception pipeline.

\section{Foundation Scale Pre-Training in 3D}
While there has been considerable advancement in the field of natural language processing and computer vision in 2D with the advent of foundation models \cite{bommasani2021foundation,radford2021clip}, the field of foundation-scale pre-training in 3D has not yet been explored \cite{guedon2023sugar,kerbl20233dgs}. This is an area where considerable advancement is yet to be achieved in the field of 3D scene understanding, especially in terms of how effectively the model is able to generalize across different environments \cite{qi2017pointnet,charles2021pointtransformer}. The area where considerable research is yet to be done in the field of foundation-scale pre-training is in how effectively self-supervised pre-training for 3D scenes can be achieved \cite{he2022mae,xie2020pointcontrast}. This could potentially include how effectively masked auto-encoding, contrastive learning, and next token prediction for 3D scenes can be achieved. Another area where considerable research is yet to be done in the field of foundation-scale pre-training is in how effectively various types of datasets, such as synthetic scenes and object-level scenes in the real world, can be integrated \cite{handa2016scenenet,dai2017scancomplete}. Another area where considerable research is yet to be done in the field of foundation-scale pre-training is in how effectively various types of model architectures can be employed for the task of pre-training \cite{qi2017pointnet++,zhou2018voxelnet}.
In order to solve these problems, different emerging research directions can be enumerated. One of these is multimodal pre-training \cite{alayrac2022flamingo,li2023blip2}. Under this, the utilization of aligned data in 2D/3D forms, such as RGB-D video, can be used for pre-training foundation models for both 2D and 3D vision \cite{wang2019rgbd,gupta2014learning}. Another emerging research direction that has been identified for solving these problems is related to different forms of tokenization. This direction can be adopted for modeling 3D scenes at different levels of abstraction \cite{yu2023pointbert,zhang2022pointmae}. Apart from this, it has been identified that the emerging research direction of diffusion pre-training is also promising, owing to the denoising objective \cite{ho2020ddpm,luo2021diffusion3d}.

\section{Generative Rendering Co-Design}

The increased connection between diffusion generative models and Gaussian Splatting has created novel opportunities for the optimization of scene representation and rendering  \cite{ho2020ddpm, rombach2022ldm, kerbl20233dgs, qian2024gsdiff}. This has created the field of generative rendering co-design and has created a number of open problems that have to be addressed. One of the problems is related to the effective usage of the iterative refinement characteristic of diffusion generative models in the optimization of the parameters of the scene representation  \cite{kerbl20233dgs, wang2024gaussiandiffusion}. Another problem is related to the development of models that can be utilized in the rendering of scenes. In this context, a number of research directions have been proposed in order to address the problems associated with the field of generative rendering co-design. One such direction is related to the optimization of Gaussians using diffusion. This is a promising direction and can be utilized in the optimization of 3D scenes using diffusion models  \cite{ho2020ddpm, rombach2022ldm}. Moreover, end-to-end hybrid training is another emerging direction that can be utilized in the optimization of the parameters of the Gaussian distribution \cite {kerbl20233dgs, lin2023magic3d}. Additionally, the optimization of geometry and appearance is gaining momentum in this context. This is particularly true in the context of emerging applications such as augmented and virtual reality.

\section{Real-Time Robotics Deployment}

Despite the fact that tremendous improvements have been made in the 3D scene understanding in the recent past, it has remained a major challenge in real-time robotics deployment, even though it has significantly improved the accuracy and reality of scenes in 3D. This is attributed to the reason that realistic scenes are very computational intensive and thus cannot be deployed in real-time robotics. However, in the future, it is necessary to develop realistic scenes that can be deployed in real-time robotics.


\section{Potential Research Directions}

Considering the current state-of-the-art developments in 3D scene completion, the following promising potential research directions may be explored. To begin with, previous approaches to scene completion were based on a volumetric scene representation~\cite{dai2017sscnet,song2017semantic}. However, such a scene representation is very computation-consuming and complicated to work with in case of fine-grain tasks~\cite{choy20163d,graham20183d}. Next, a major problem concerning deep learning model application for scene completion is domain generalization. Thus, most scene completion approaches rely on artificial data generation, which results in lack of domain generalization in particular in terms of the difference between indoor and outdoor environments~\cite{handa2016scenenet,chen2021scnet}. The recent improvement of scene completion performance comes from the usage of different cues (RGB images, depth maps, semantics, etc.) in understanding the scene structure~\cite{jaritz2019multi,chen2020multiview}. Moreover, development of efficient models capable of performing in real-time is another promising research direction since robots are becoming increasingly popular~\cite{tang2022efficient}.
\subsection{Summary of Future Research Directions}
\label{sec:future_directions_summary}

Table~\ref{tab:future_directions} synthesizes the key future research directions with estimated timeframes and impact, providing a roadmap for the next generation of 3D scene completion systems.

\begin{table}
\centering
\caption{Key future research directions for 3D scene completion.}
\label{tab:future_directions}
\begin{tabular}{p{3.2cm}p{2.2cm}p{3.2cm}p{4cm}}
\toprule
\textbf{Research Direction} & \textbf{Timeframe} & \textbf{Expected Impact} & \textbf{Key Challenges} \\
\midrule
Foundation 3D Models & 3--5 years & High – Enable few-shot completion & Data scale, architecture design \\
Real-time Gaussian-Diffusion Hybrids & 1--2 years & High – Deployable AR/robotics & Inference speed, training stability \\
Dynamic Scene Completion & 3--4 years & Medium – Video-rate 4D understanding & Representation, temporal consistency \\
Uncertainty Quantification & 2--3 years & Critical – Safety certification & Calibration, evaluation metrics \\
Multi-modal Foundation Models & 4--6 years & High – Language-guided completion & Alignment, computational cost \\
Green AI for 3D & Ongoing & Environmental – Sustainable research & Efficiency metrics, model compression \\
\bottomrule
\end{tabular}
\end{table}

\section{Reproducibility and Benchmarking}
Reproducibility is one of the most important factors that are taken care of in the recent research studies related to machine learning. There are a lot of research studies that have already proved the enhancement of the machine learning model. However, there is no standard setting in those studies. Hence, reproducibility is one of the most important factors that have to be considered.

\section{Conclusion}
\label{sec:conclusion}
The insights that would emerge for the readers based on the observed trends in modern vision architecture can be analyzed in the context of representation architecture, architectural innovations, data, evaluation, and application of the trends. Some of the insights identified based on the trends that are emerging in recent years are as follows. To begin with, one of the important trends is that of transition from convolutional architecture to sequence and attention architecture. Second, there have also been cases of using hybrid architectures where several vision architectures have been combined together. Third, there have been advancements made with respect to state-space vision architectures like Vision Mamba.


\begin{backmatter}

\section*{Acknowledgements}
The authors gratefully acknowledge the support from the Deanship of Scientific Research and Graduate Studies at King Khalid University (Grant No. RGP2/104/45) and Yayasan Universiti Teknologi PETRONAS (YUTP-PRG 015PBC-028).

\section*{Availability of data and materials}
Not applicable.

\section*{Ethics approval and consent to participate}
Not applicable.

\section*{Competing interests}
The authors declare that they have no competing interests.

\section*{Consent for publication}
Not applicable.

\section*{Authors' contributions}
A.K. conceptualized the study, designed the methodology, conducted the systematic review and meta-analysis, performed the literature search and data extraction, wrote the main manuscript text, and prepared all figures, tables, and supplementary materials. S.J.A. reviewed, revised, and edited the manuscript for intellectual content,  secured the funding, provided project administration and supervision, and facilitated the resources for the study. M.M.E.E. secured the funding. All authors reviewed the final manuscript and approved its submission.

\bibliographystyle{bmc-mathphys} 
\bibliography{bmc_article}     
\end{backmatter}
\end{document}